\documentclass[11pt]{article}
\usepackage[a4paper,top=3cm,bottom=2cm,left=2cm,right=2cm,marginparwidth=1.75cm]{geometry}

\usepackage[utf8]{inputenc} 
\usepackage[T1]{fontenc}
\usepackage[numbers]{natbib}
\usepackage[colorlinks=True,citecolor=blue,urlcolor=blue,pagebackref=true,backref=true]
{hyperref}
\renewcommand*{\backrefalt}[4]{%
    \ifcase #1 \footnotesize{(Not cited.)}%
    \or        \footnotesize{(Cited on page~#2.)}%
    \else      \footnotesize{(Cited on pages~#2.)}%
    \fi}

\usepackage[none]{hyphenat}
\usepackage{breakcites}
\usepackage[utf8]{inputenc} 
\usepackage[T1]{fontenc}    
\usepackage{hyperref}       
\hypersetup{
  colorlinks = true,
  urlcolor = blue,
  linkcolor = blue,
  citecolor = blue
}
\usepackage{booktabs}       
\usepackage{amsfonts}       
\usepackage{nicefrac}       
\usepackage{microtype}      
\usepackage{xcolor}         
\usepackage{caption}
\usepackage{subcaption}
\usepackage{float}
\usepackage{amsmath,amsthm}
\usepackage{amssymb}
\usepackage{mathtools}
\usepackage{soul}
\usepackage{bm}
\usepackage{enumitem}
\usepackage{multirow}
\usepackage{wrapfig}
\usepackage{scalerel}
\allowdisplaybreaks
\usepackage{dsfont}

\usepackage{xcolor}
\usepackage{thmtools,thm-restate}
\usepackage{cleveref}

\usepackage{dsfont}

\usepackage[noend]{algpseudocode}
\usepackage{algorithm}
 \newtheorem{theorem}{Theorem}
 \newtheorem{definition}{Definition}
 \newtheorem{lemma}{Lemma}

 \newtheorem{proposition}{Proposition}

\newcommand{\bX}{\boldsymbol{X}}
\newcommand{\bY}{\boldsymbol{Y}}

\newcommand{\bU}{\mathbf{U}}
\newcommand{\bV}{\mathbf{V}}

\newcommand{\subG}{\,\mathsf{subG}\,}

\newcommand{\MNN}{\texttt{MNN}}
\newcommand{\Impute}{\texttt{Impute}}

\newcommand{\USVT}{\texttt{USVT}}
\newcommand{\SNN}{\texttt{SNN}}

\newcommand{\E}{{\rm I}\kern-0.18em{\rm E}}

\newcommand{\cC}{\mathcal{C}}

\newcommand{\cE}{\mathcal{E}}
\newcommand{\cS}{\mathcal{S}}
\newcommand{\cD}{\mathcal{D}}
\newcommand{\cU}{\mathcal{U}}
\newcommand{\cV}{\mathcal{V}}
\newcommand{\cI}{\mathcal{I}}
\newcommand{\cJ}{\mathcal{J}}

\newcommand{\cB}{\mathcal{B}}
\newcommand{\cY}{\mathcal{Y}}

\newcommand{\Pb}{\mathbb{P}}

\newcommand{\Reals}{\mathbb{R}}

\newcommand{\NN}{\mathbb{N}}

\DeclarePairedDelimiter{\norm}{\lVert}{\rVert}
\DeclarePairedDelimiter{\abs}{\lvert}{\rvert}
\DeclarePairedDelimiter{\parens}{\lparen}{\rparen}
\DeclarePairedDelimiter{\sqbrac}{[}{]}

\newcommand{\enorm}[1]{\norm*{#1}_2}

\newcommand{\opnorm}[1]{\norm*{#1}_{\operatorname{op}}}
\newcommand{\fnorm}[1]{\norm*{#1}_{F}}
\newcommand{\maxnorm}[1]{\norm*{#1}_{\operatorname{max}}}

\newcommand{\bz}{\mathbf{z}}

\newcommand{\bv}{\mathbf{v}}
\newcommand{\bu}{\mathbf{u}}
\newcommand{\bx}{\mathbf{x}}
\newcommand{\by}{\mathbf{y}}

\newcommand{\bq}{\mathbf{q}}
\newcommand{\bw}{\mathbf{w}}

\newcommand{\ba}{\mathbf{a}}

\newcommand{\bfm}{\mathbf{m}}

\newcommand{\bfX}{\mathbf{X}}
\newcommand{\bfA}{\mathbf{A}}
\newcommand{\bfP}{\mathbf{P}}
\newcommand{\bfE}{\mathbf{E}}
\newcommand{\bfU}{\mathbf{U}}
\newcommand{\bfV}{\mathbf{V}}
\newcommand{\bfH}{\mathbf{H}}
\newcommand{\bfM}{\mathbf{M}}
\newcommand{\bfY}{\mathbf{Y}}
\newcommand{\bfI}{\mathbf{I}}

\newcommand{\bfQ}{\mathbf{Q}}
\newcommand{\bfR}{\mathbf{R}}

\newcommand{\bfSigma}{\mathbf{\Sigma}}

\newcommand{\prob}[1]{\Pb\left[ #1 \right]}

\newcommand{\dnoise}{\delta_{\operatorname{noise}}}
\newcommand{\ddist}{\delta_{\operatorname{dist}}}

\newcommand{\du}{d^\bu}
\newcommand{\hatdu}{\hat{d}^\bu}

\newcommand{\hatdv}{\hat{d}^\bv}

\newcommand{\EE}{\mathbb{E}}
\newcommand{\PP}{\mathbb{P}}
\newcommand{\RR}{\mathbb{R}}

\newcommand{\indicator}{\mathds{1}}
\newcommand{\op}{{\textup{op}}}
\newcommand{\F}{\textup{F}}

\title{Exploiting Observation Bias to Improve Matrix Completion}
\author{Yassir Jedra \\ MIT \\ jedra@mit.edu \and  Sean Mann \\ MIT \\ seanmann@mit.edu \and  Charlotte Park \\ MIT \\ cispark@mit.edu \and Devavrat Shah \\ MIT \\ devavrat@mit.edu}

\date{}

\begin{document}
\maketitle

\setlength{\parindent}{0cm}

\begin{abstract}
We consider a variant of matrix completion where entries are revealed in a biased manner. We wish to understand the extent to which such bias can be exploited in improving predictions. Towards that, we propose a natural model where the observation pattern and outcome of interest are driven by the same set of underlying latent (or unobserved) factors. We devise Mask Nearest Neighbor (MNN), a novel two-stage matrix completion algorithm: first, it recovers (distances between) the latent factors by utilizing matrix estimation for the fully observed noisy binary matrix, corresponding to the observation pattern; second, it utilizes the recovered latent factors as features and sparsely observed noisy outcomes as labels to perform non-parametric supervised learning. Our analysis reveals that MNN enjoys \emph{entry-wise} finite-sample error rates that are competitive with corresponding supervised learning parametric rates. Despite not having access to the latent factors and dealing with biased observations, MNN exhibits such competitive performance via only exploiting the shared information between the bias and outcomes. Finally, through empirical evaluation using a real-world dataset, we find that with MNN, the estimates have 28x smaller mean squared error compared to traditional matrix completion methods, suggesting the utility of the model and method proposed in this work.   
\end{abstract}

\maketitle


\maketitle

\section{Introduction}

Matrix completion is a fundamental task in modern machine learning that has garnered widespread interest across many societal and engineering domains \citep{hazewinkel2022sensitivity, steck2010training, abadie2010synthetic, davenport2016overview}. In matrix completion, the goal is to estimate a matrix from partial and noisy observations of its entries.  Specifically, we aim to recover an $m \times n$ matrix $\bfX = (x_{ij})_{(i,j)\in [m]\times[n]}$ with access to noisy observations of only a small subset of its entries. This subset of observed entries can be characterized by a mask or what we refer to as the observation matrix $\bfA = (a_{ij})_{i,j\in [m] \times [n]}$ taking values in $\lbrace 0, 1\rbrace$, where $a_{ij} = 1$ if a noisy version $y_{ij}$ of $x_{ij}$ is observed, and $0$ otherwise (entry $(i,j)$ is missing) for $(i, j) \in [m] \times [n]$. 

\paragraph{The MCAR framework.} Much of the existing literature on matrix completion has been focused on the so-called missing completely at random (MCAR) model. In this framework, all entries are revealed independently $a_{ij} \sim \mathrm{Ber}(p)$, i.e. $a_{ij}=1$ with probability $p$ and $0$ otherwise for $p \in (0,1]$. Some early prominent works under the MCAR framework have been based on convex optimization and spectral methods \citep{candes2009power, recht2011simpler, cai2010singular, keshavan2009matrix, keshavan2012matrix, chatterjee2015matrix, chen2020noisy}. Specifically, the work of \cite{chatterjee2015matrix} proposed a simple estimation procedure known as Universal Singular Value Thresholding (USVT) that achieves the minimax error rate up to a constant factor under the MCAR framework. In another line of work, non-pararmetric methods based on collaborative filtering have been shown to achieve consistent estimation, notably \cite{li2019nearest}, and \cite{borgs2021iterative}. 

\paragraph{The MNAR framework.} Despite this extensive literature, the MCAR model is typically not applicable in practice, as there are \emph{unobserved covariates} that are influencing the observation pattern as well as outcome of interest. Indeed, in many real world applications, such as public policy evaluation \citep{abadie2010synthetic}, or recommender systems \citep{hazewinkel2022sensitivity}, the observation pattern encodes information that can also help estimate the outcome of interest. As an example, consider a movies recommender system where the rows of $\bfX$ correspond to users, the columns correspond to movies, and an entry $x_{ij}$ corresponds to a user $i$'s preference of movie $j$. In this scenario, we wish to understand how a user would rate a movie they have not seen, allowing us to provide the user with recommendations for new movies they are likely to enjoy. We note, however, that users are less likely to view movies in genres they do not like, meaning that they are less likely to rate these movies. Clearly the observation pattern for such a matrix is not uniform at random as users’ genre preferences affect which movies they give ratings for. Hence, in order to understand the practical conditions under which matrix completion can be performed, we must focus on settings where observations are missing not at random (MNAR) \citep{little2019statistical}. Furthermore, we note that the presence of such bias in observations is an example of the classical problem of \emph{confounding}, meaning, the factors driving the reasons for not observing an outcome are also influencing the outcome itself. In the setting of matrix completion, however, these confounding factors are \emph{unobserved}. This is exactly what has driven the recent interest in the MNAR framework. 

\medskip

\paragraph{The Question Of Interest.} 
While there is exciting recent progress in the MNAR framework (see Section \ref{ssec:prior-mnar} for details), the following question remains unaddressed: 

\begin{center}
    {\em Can bias in observations, due to the presence of latent factors that influence both the observation pattern and the outcomes, aid in estimating outcomes of interest?}
\end{center}

\subsection{Prior work on the MNAR framework}\label{ssec:prior-mnar} Many efforts have been made recently towards addressing the challenges posed by matrix completion in the MNAR framework \citep{ma2019missing, agarwal2021causal, bhattacharya2022matrix, yang2021tenips, liang2016modeling, schnabel2016recommendations, wang2020recsys, sportisse2020imputation, sportisse2020estimation, abadie2024doubly}. These diverse contributions bring  unique perspectives on the MNAR framework that are reflected in the differing settings they consider which is indicative of the complexity of the framework itself. Among these studies, we further discuss three works, namely those by \cite{ma2019missing, bhattacharya2022matrix, agarwal2021causal},  that are representative of prior works. 
 
\paragraph{Confounding-agnostic MNAR models.}  \cite{ma2019missing} consider an MNAR model where the so-called {\em propensity score matrix} $\bfP = (p_{ij})_{(i,j)\in [m]\times[n]}= \EE[\bfA]$\footnote{For $(i,j) \in [m] \times [n]$, 
$p_{ij}$ is the probability of observing the entry $(i,j)$ and is denoted here as propensity score in the language of Causal Inference.}  is assumed to have low nuclear norm and is estimated directly. This estimate is then used to recover the outcome matrix by leveraging the classical technique of inverse propensity weighing (IPW) \citep{imbens2015causal}. Unfortunately, the use of such technique requires the propensity scores $p_{ij}$ to be uniformly bounded away from $0$ in order for their methods to be provably consistent in a mean squared error (MSE) sense. Because they follow a similar approach, other works such as \cite{yang2021tenips, liang2016modeling, schnabel2016recommendations, wang2020recsys, abadie2024doubly} suffer from the same limitation. In contrast, it is worth noting that our work evades such limitation.

\medskip 

\cite{bhattacharya2022matrix} consider a self-masking MNAR model where the propensity scores are a  scalar function of their corresponding outcomes, i.e., $p_{ij} = \omega(x_{ij})$ with $\omega$ being the scalar function a priori unknown. Under such model, the authors propose modified versions of classic spectral-based methods like USVT and establish their consistency in an MSE sense. \cite{sportisse2020imputation, sportisse2020estimation} follow similar modelling choices and also propose spectral-based methods that are also consistent in the same sense. This collection of works consider a model that can be viewed as a special instance of the model considered in the present work as we shall see in subsequent sections.  

\medskip

None of the considered MNAR models in the aforementioned works account for the presence of confounding in matrix completion, i.e., namely the presence of \emph{unobserved confounding factors} that can influence both the missiness pattern characterized by $\bfP$ and outcome of interest $\bfX$.  Consequently, the methods these works propose are agnostic to the presence of confounding, thus leading these works to treat the observation bias as a disadvantage that needs mitigating rather than exploiting as we do. Furthermore, we provide much sharper performance guarantees (entry-wise error bound) compared to MSE. It is worth noting that entry-wise guarantees are much harder to obtain even in the MCAR framework and only recently such results have emerged, e.g. see \cite{abbe2019entrywise, chen2021spectral, borgs2021iterative} and references therein. 

\paragraph{Confounding-aware MNAR models.} In the nascent literature of matrix completion under the MNAR framework, models that account for the presence of confounding are rather scarce. One work that stands out in this scarce literature is that of  \cite{agarwal2021causal}, where they consider an MNAR model with shared information between the observation pattern and outcomes that is determined by a set of \emph{unobserved confounding} or latent factors.
Specifically, the outcomes have a low-rank representation with respect to the latent factors while the observation pattern has generic dependence on the latent factors. They establish that despite generic dependence between observation pattern and latent factors, as long as there is certain ``rectangular'' pattern that is part of observed entries, their proposed algorithm Synthetic Nearest Neighbours (SNN) can provide \emph{entry-wise} performance guarantees. In that sense, the goal of \cite{agarwal2021causal} has been to show that it is feasible to estimate outcomes of interest even when observation pattern is confounded. However, this work stops short of utilizing the {\em shared} information between observation pattern and outcomes to {\em improve} the matrix estimation. And this is exactly the primary motivation for this work. 

\medskip

Indeed, devising matrix completion methods that take advantage of such information is an important aspect that has not been addressed. Addressing it is exactly what our work aims to do, thus already distinguishing our contributions from all prior work on the MNAR framework.

\subsection{Contributions}

We consider that entries are revealed independently, i.e. $a_{ij} \sim \mathrm{Ber}(p_{ij}), i,j \in [n]$ are mutually independent\footnote{In this work, for ease of exposition, we use $n\times n$ matrix. As the reader will notice, the methods and analysis naturally extend for $n \times m$ matrix with $m \neq n$.}, where the observation probabilities or propensity scores, $\bfP = (p_{i,j})_{(i,j) \in [n]}$ may take values in $[0,1]$ and vary across rows and columns. If $a_{ij} = 1$, then $y_{ij}$ is noisy observation of $x_{ij}$ revealed independently. Starting with this framework, our contributions are listed below:

\begin{itemize}
    \item[(a)] \emph{A novel confounding-aware MNAR model.} To begin with, we propose a novel model within the MNAR framework for matrix completion where we consider that the observation probabilities $\bfP$ and outcome of interest $\bfX$ are determined by the same set of latent factors $(\bu_i)_{i \in [n]}, (\bv_j)_{j \in [n]}$, which take values in the unit sphere $\cS^{d-1}$ for $d > 1$. More specifically, we consider for all $i,j \in [n]$, that $p_{ij} = n^{-\beta} (1 + \bu_i^\top \bv_j)/2$ for some $\beta > 0$, and $x_{ij} = f(\bu_i, \bv_j)$ for some a priori unknown Lipschitz function $f$. We observe noisy outcomes $\bfY = (y_{ij})_{i,j \in [n]}$ where $y_{ij} = \EE[x_{ij}]$ if $a_{ij} = 1$, otherwise  $y_{ij} = \star$, to mean that the entry is missing. We refer the reader to Section \ref{sec:problem} for a formal model description and the rationale behind it.
    \item[(b)] \emph{A confounding-aware matrix completion method with entry-wise performance guarantees.} Next, we introduce the Mask Nearest Neighbors ($\MNN$) to estimate $\bfX$ using the mask matrix $\bfA$ and the noisy observation matrix $\bfY$. The algorithm proceeds in two-stages: (i) estimate distances between the latent vectors using $\bfA$; (ii) it utilizes $\bfY$ along with distance estimates between latent factors to apply non-parametric supervised learning to produce estimate of $\bfX$. We show that $\MNN$ enjoys entry-wise error rates that scale as $\tilde{O}(n^{-(2-\beta)/(2d)})$ provided that $\beta < 1/4$ (see Theorem \ref{thm:prediction}). 
    \item[(c)] \emph{Comparing with lower bounds in supervised learning.} In the traditional supervised learning setup, the goal is to learn the function $f(\cdot, \cdot)$ using $N$ i.i.d. observations sampled as per a distribution over the domain $\cS^{d-1} \times \cS^{d-1}$. It is well established that under reasonable conditions, to obtain consistent estimation of the function, it requires $\tilde{\Omega}(N^{-1/(2d)})$ (see Lemma \ref{lem:prediction-param-rate}) samples. In our setup, on average $N = \Theta(n^{2 - \beta})$ observations are available. Therefore, without additional structure it must require $\tilde{\Omega}(N^{-1/(2d)}) = \tilde{\Omega}(n^{-(2-\beta)/(2d)})$ observations for consistent estimation of $f$ or $\bfX$. Indeed, such lower bound would assume that we {\em observe} the actual values of latent factors. As noted earlier, we are able to obtain consistent estimation of $\bfX$ with effectively similar number of observations, i.e. $\tilde{O}(n^{-(2-\beta)/(2d)})$ despite not observing latent factors due to the {\em factorization} of $f$. Indeed, the precise number of observations required is less than the lower bound and it is not violating it because of the factorization structure of $f$. The purpose of this discussion is to simply contextualise the non-triviality of our results.
    \item[(d)] \emph{Empirical validation on a real-world dataset.} To establish relevance of the model and subsequently resulting algorithm, $\MNN$, we consider a real-world setting. Specifically, we utilize more than a million interaction data points of users on the Glance\footnote{Please note that the dataset we use is not publicly available. However, it can be provided by Glance upon request for verification purposes only. For further information about Glance, we refer the reader to their website: \url{https://glance.com/us}.} platform -- a smart lock-screen that aims to personalize user experience through recommending dynamic lock screens (also called {\em glances}). We find that $\MNN$ can improve the mean-squared error by 28x compared to a standard matrix completion method (see Table \ref{table:real_performance}). We also report empirical performance using a synthetic dataset to discuss nuanced properties of $\MNN$ that are not necessarily captured by theoretical results (see Section \ref{sec:experiments}).
\end{itemize}

\subsection{Notation} We recall that $\cS^{d-1}$ denotes the unit sphere in $\RR^d$. We write $[n]$ to mean $\lbrace 1, \dots, n\rbrace$. For a given parameter $p \in [0,1]$, $\mathrm{Ber}(p)$ denotes the Bernoulli distribution with parameter $p$. For any  $a, b \in \RR$, we write $a \wedge b = \min(a, b)$ and $a \vee b = \max(a,b)$. For any given vector $\bx \in \RR^d$, we denote its euclidean by $\norm{\bx}$. For a given $m\times n$ matrix $\bfM$, for any subsets $\cI \subseteq [n]$, $\cJ \subseteq [m]$, we denote $\bfM_{\cI, \cJ}$ the sub-matrix of $\bfM$ that contains only the entries $\cI \times \cJ$. We further use $\bfM_{i,:}$ (resp. $\bfM_{:,j}$) to denote the $i^{\text{th}}$ row (resp. $j^{\text{th}}$ column) of $\bfM$. We denote the max norm of $\bfM$ by $\Vert \bfM \Vert_{\max} = \max_{(i,j) \in [m] \times [n]} \vert \bfM_{i, j} \vert$, its operator norm by $\Vert \bfM \Vert_\op$, its Frobenius norm by $\Vert \bfM \Vert_\F$, and its two-to-infinity norm by $\Vert \bfM\Vert_{2 \to \infty} = \max_{i \in [m]}\Vert \bfM_{i,:}\Vert$. The pseudo-inverse of the matrix $\bfM$ is denoted by $\bfM^\dagger$.

\section{Model Description} \label{sec:problem} 

In this section, we present a natural MNAR model that takes into account the presence of unobserved confounding. We provide the reasoning 
behind this being the natural model.

\subsection{The MNAR model}\label{subsec:mnar model}

We consider a recommender system setting where the number of users and number of items is equal to $n$. Note that we only make this choice for ease of exposition, but otherwise our results easily extend to the case where the set of items and users are not of equal size.  

\paragraph{Latent factor model.} We posit that users (rows) and items (columns) can be fully described by $d$-dimensional unit vectors where $d \ll n$. More specifically, each user $i$ (resp. item $j$) is associated with a unit vector $\bu_i \in \cS^{d-1}$ (resp. $\bv_j \in \cS^{d-1}$), which we assume to be distributed independently and uniformly at random over the unit sphere. We will refer to $(\bu_i)_{i \in [n]}$ (resp. $(\bv_j)_{j \in [n]}$) as the \emph{user} (resp. \emph{item}) \emph{latent factors}. Such latent factors determine both the probability of observation as well as the outcome of interest.

\paragraph{Outcomes.} The outcomes of interest may be interpreted as affinity scores between user-item pairs. In our model, we denote the outcome of each user-item pair $(i,j) \in [n]\times[n]$ by $x_{ij}$, and  consider that it may be expressed as a sum of $d$ separable functions: 
\begin{align}\label{eq:rep-outcome}
    x_{ij} = f(\bu_i, \bv_j) = \sum_{\ell = 1}^d  \theta_\ell(\bu_i) \phi_\ell(\bv_j),
\end{align}
where the mappings $\theta(\cdot) = (\theta_1(\cdot), \dots, \theta_d(\cdot))$ and $\phi(\cdot) = (\phi_1(\cdot), \dots, \phi_d(\cdot))$ are defined from $\cS^{d-1}$ to $[-B, B]^{d}$ for some $B >0$, and are assumed to be $L$-lipschitz. As such, the outcomes, despite forming a low-rank matrix, are not simply bilinear forms of the \emph{user} and \emph{item} latent factors as is traditionally the case in matrix completion.

\paragraph{Observation model.}

 Our setup differs from the classical matrix completion literature in that entries are \emph{not} randomly revealed with uniform probability. Instead, the observation probability of a given user-item pair $(i,j)$ is defined as:
\begin{equation}\label{eq:prob-obs}
    p_{ij} = \frac{\rho_n}{2} (\bu_i^\top \bv_j + 1),
\end{equation}
where $\rho_n \in (0,1]$ is a \emph{sparsity factor} that may vary with $n$. This means that the observation pattern induced is very closely related to the \emph{random dot product graph} construction \citep{athreya2018statistical}, and  \emph{geometric random graphs} \citep{duchemin2023random}. This captures the selection bias in recommender systems \citep{ma2019missing}. Let $a_{ij} \sim \mathrm{Ber}(p_{ij})$, independent across user-item pairs $i, j \in [n]$: 
$a_{ij} = 1$ means that the entry $(i,j)$ is revealed, and we observe a \emph{noise-corrupted} outcome $y_{ij}$. Specifically, 
\begin{align}
    y_{ij} = \begin{cases}
        x_{ij} + \xi_{ij} & \textrm{if } a_{ij} = 1,\\
        \star & \textrm{otherwise}
    \end{cases}
\end{align}
where $\star$ indicates that an entry is not revealed, and $(\xi_{ij})_{i,j \in [n]}$ are 
independent zero-mean and $\sigma^2$-sub-gaussian random variables.

\paragraph{Objective.} Given the observation pattern $(a_{ij})_{i,j \in [n]}$, and observations $\cD = \lbrace y_{ij}:  a_{ij} = 1,   i,j \in [n]\rbrace$, produce an estimate $\widehat{\bfX} = (\hat{x}_{ij})_{(i,j)\in [n]}$ of $\bfX = (x_{ij})_{i,j \in [n]}$
so that the entry-wise prediction error $\vert \hat{x}_{ij} - x_{ij}\vert$, for all $i,j \in [n]$ is as small as possible, with high probability (i.e. probability going to $1$ as $n\to\infty$).  

\medskip

\paragraph{Further notation.}  In what follows, we will often prefer matrix notation to ease readability. Thus, we denote $\bU^\top = \begin{bmatrix} \bu_1 & \cdots & \bu_n\end{bmatrix}$, $\bV^\top = \begin{bmatrix} \bv_1 & \cdots & \bv_n\end{bmatrix}$ in $\RR^{d \times n}$,  $\bfX = (x_{ij})_{i,j\in [n]}$, $\bfP = (p_{ij})_{i,j\in [n]}$, $\bfA = (a_{ij})_{i,j \in [n]}$  and $\bfY = (y_{ij})_{i,j\in [n]}$ in $\RR^{n \times n}$. 

\subsection{Why this model makes sense?}

\paragraph{Presence of confounding.} The motivation behind the model described in \textsection \ref{subsec:mnar model} is precisely to capture a phenomenon that often occurs in practice, namely that of \emph{confounding}. An example of this is the case of the Glance smart lock-screens, where the rows of $\bfX$ correspond to users, the columns correspond to individual glances and an entry $x_{ij}$ corresponds to the time user $i$ spends interacting with glance $j$.  We note here that a user not interested in sports is unlikely to spend time viewing a glance recounting a recent basketball game. Moreover, based on the behavior of this user, the recommendation algorithm is unlikely to recommend glances about sports. Hence, information about the users' preferences can be extracted both from their engagement time and which lock screens they are shown. This shared information corresponds to common latent features that affect both the observation pattern and the outcome in our model.

\paragraph{A generic confounding-aware MNAR framework.} Formally, we may consider that the common features as discussed above for a user-item pair $(i,j)$ may be denoted by the two $d$-dimensional vectors $\mu_i$ and $\nu_j$. Then, as posited before, $\mu_i$ and $\nu_j$ influence both the the probability $p_{ij}$, corresponding to observing the outcome of interest for the user-item pair $(i,j)$, as well as the outcome of interest itself $x_{ij}$:   
\begin{align}
    p_{ij} & = g(\mu_i, \nu_j), \\
    x_{ij} & = h(\mu_i, \nu_j),  
\end{align}
where $g$ (resp. $h$) is a function that maps the common factors to a value in $[0,1]$ (resp. in some bounded domain of possible outcome values). Without further assumptions on $g$ and $h$, the task of matrix completion may be unsolvable. It therefore natural to assume such functions to be \emph{nice} in some appropriate sense. The model we described here is somewhat general and is of independent interest. We refer to a concurrent work \citep{jalan2024transfer} where a similar shared latent factored model was considered in the context of transfer learning between biological networks.

\paragraph{The low-rank hypothesis.} It is reasonable to consider that $g$ and $h$ are \emph{nice} functions in the sense that they are bounded and Lipschitz. In this case, we may use a classical functional decomposition result \citep{conway2019course}. We present below a variant of such decomposition (see Theorem 1 in \citep{shahRL}):      
\begin{proposition}\label{thm:f}
    Let $g: \cS^{d-1}\times \cS^{d-1} \mapsto \RR$ be bounded and $L$-Lipschitz with respect to the $1$-product metric in $\cS^{d-1} \times \cS^{d-1}$. Then, there exists a non-increasing sequence $(\sigma_k)_{k \ge 1}$ with $\sum_{k=1}^\infty \sigma_k^2 < \infty$, and orthogonal families of functions $\lbrace \alpha_k \in L^2(\cS^{d-1}): k \in \NN^\star \rbrace$, and  $\lbrace \gamma_k \in L^2(\cS^{d-1}): k \in \NN^\star \rbrace$ such that,  for all $\bz, \bw \in \cS^{d-1}$,
    \begin{align}
         g(\bz, \bw) = \sum_{k = 1}^\infty \sigma_k \alpha_k(\bz)\gamma_k(\bw). 
    \end{align}
\end{proposition}
Using Proposition \ref{thm:f}, it is reasonable to assume that both $g$ and $h$ admit a low-rank representation, say of rank $d$ for simplicity, where we write 
$ p_{ij}  = g(\mu_i, \nu_j) \approx \sum_{\ell = 1}^d a_{\ell}(\mu_i) b_{\ell}(\nu_j)$ and $ x_{ij}  = h(\mu_i, \nu_j) \approx \sum_{\ell = 1}^d \alpha_{\ell}(\mu_i) \beta_\ell(\nu_j),
$
where 
$\alpha(\cdot), \beta(\cdot), a(\cdot), b(\cdot)$ are mappings from $\cS^{d-1}$ to $\RR^{d}$. Thus, if additionally $a(\cdot)$ and $b(\cdot)$ are invertible, then we may define $\theta = \alpha \circ a^{-1}$ and $\phi = \beta \circ b^{-1}$, and finally obtain (with appropriately
defined $\bu_i, \bv_j$ with $i, j \in [n]$)
\begin{align}
    p_{ij} & \approx \bu_i^\top \bv_j, \label{eq:p_deriv} \\
    x_{ij} & \approx \sum_{\ell = 1}^d \theta_\ell(\bu_i)  \phi_\ell(\bv_j).\label{eq:x_deriv} 
\end{align}
We clearly see the similarity between \eqref{eq:rep-outcome} (resp. \eqref{eq:prob-obs}) and \eqref{eq:x_deriv} (resp. \eqref{eq:p_deriv}). Therefore, we see the model we described in \textsection \ref{subsec:mnar model} as a natural and fairly general instance of a an MNAR model that accounts for the presence of confounding.

\section{Mask Nearest Neighbor (MNN)}\label{section:alg}

In this section, we present the Mask Nearest Neighbors ($\MNN$) algorithm, which gets its name from its two stages: (i) {\em Learning Distances From Masks}: estimate distances between the latent factors associated with users, items 
using mask matrix $\bfA$, and using them to cluster users, items; (ii) {\em  Estimation using Nearest Neighbor}: de-noising observed values and impute missing values in the outcome matrix via non-parametric nearest neighbor supervised learning.

\subsection{Distance Estimation, Clustering}

\emph{Distance estimation.} The first step of the algorithm is to estimate the user-user and item-item distances in the latent space, denoted $d_{ij}^\bu = \Vert \bu_i - \bu_{j} \Vert_2$, and $d_{ij}^\bv = \Vert \bv_i - \bv_{j} \Vert_2$, for all $i,j \in [n]$, respectively. 
To that end, we start by constructing the centered matrix of observations $\widetilde{\bfA} = \bfA - (\rho_n/2) \bf1 \bf1^\top$, which we then project onto the space of rank $d$ matrices. More precisely, given the singular value decomposition $\widetilde{\bfA} = \sum_{i=1}^n \sigma_i \bq_i \bw_i^\top$ we construct an estimate of $\widetilde{\bfP} = \EE[\widetilde{\bfA}]$ as follows: 
\begin{align}
    \widehat{\widetilde{\bfP}} = \sum_{i=1}^d \sigma_i \bq_i \bw_i^\top,  
\end{align}
where $\sigma_1 \geq \dots \geq \sigma_n$ are the singular values of $\widetilde{\bfA}$, and $\bq_i$ and $\bw_i$ are the corresponding left and right singular vectors. Then,
for users (resp. items) $i$ and $j$ we can then define the \emph{distance estimate}
\begin{align}\label{eq:distance-estimation}
     \hatdu_{ij} & = \frac{2}{\rho_n} \sqrt{\frac{d}{n}}  \enorm{\widehat{\widetilde{\bfP}}_{i, :} - \widehat{\widetilde{\bfP}}_{j, :}}, \\ 
    \hatdv_{ij}  & = \frac{2}{\rho_n} \sqrt{\frac{d}{n}}   \enorm{\widehat{\widetilde{\bfP}}_{:, i} - \widehat{\widetilde{\bfP}}_{:,j }}.
\end{align}
In above, $\widetilde{\bfP}_{i, :}$ (resp. $\widetilde{\bfP}_{:, i}$) correspond to $i$th row (resp. $i$th column) of $\widetilde{\bfP}$.

\medskip 

 \emph{Clustering.} The goal is to cluster users (resp. items) using their distance estimates so that any two users (resp. items) that belong to the same cluster are within distance $\varepsilon_n > 0$, a hyper-parameter of choice. Towards that, first we identify {\em central users}, denoted as $\cU^\star \subset [n]$, 
that are farther apart from each other. To identify $\cU^\star$, start with it to be empty set. 
We repeatedly add an arbitrary user to the set if its estimated distance from all user in $\cU^\star$ is 
at least $6\varepsilon_n$. Stop when no more users can be added. Similarly construct $\cV^\star$ for items. 

\medskip 

We shall create cluster of users around each central user in $\cU^\star$. To that end, for each $i \in [n] \backslash \cU^\star$, 
associate it to the closest of the users in $\cU^\star$ as per the distance estimated. Thus, each $i \in \cU^\star$ corresponds to a central
user, and let $\cU_i \subset [n]$ be the users associated with it. Thus we have a partition or clusters of the users 
$\bigcup_{i \in \cU^\star} \cU_i = [n]$. Similarly construct the item partition $\bigcup_{j \in \cV^\star} \cV_j$. Note that we abuse
notation here by indexing the partitions by the elements of $\cU^\star, \cV^\star$ to avoid overload of notation.

\medskip 

The observations can then be partitioned as $\cD = \bigcup_{i \in \cU^\star} \bigcup_{j \in \cV^\star} \{y_{k\ell}: (k, \ell) \in \Omega_{ij}\}$, where $\Omega_{ij} = \{(k, \ell): a_{k\ell} = 1, k \in \cU_i, \ell \in \cV_j\}$ is the set of observed outcomes 
between users in cluster $i$ and items in cluster $j$.

\begin{figure}
    \centering
    \includegraphics[width=1\linewidth]{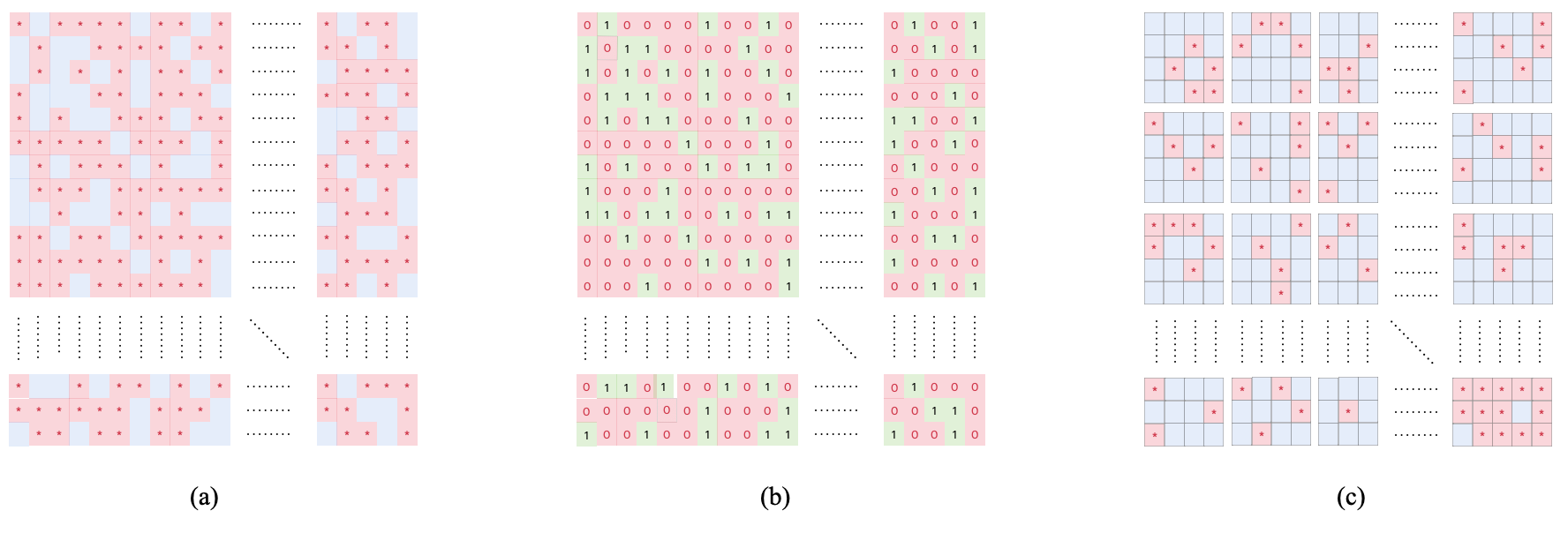}
    \caption{\emph{Visual depiction of the first stage of MNN.} The matrix of outcomes $\bfY$ is illustrated by figure (a). From the outcomes matrix $\bfY$, we extract the matrix $\bfA$ which is fully observed and illustrated by figure (b). The matrix $\bfA$ is then used to learn  distances between the shared latent user factors (resp. latent item factors), then use these distances to cluster users (resp. items). Using these clusters, we may rearrange the rows and columns of the outcome matrix as illustrated by figure (c), where each block correspond to the Cartesian product of a cluster of users and a cluster of items.}
    \label{fig:enter-label}
\end{figure}

\subsection{Outcome Prediction}

\emph{Denoising.} Define the \emph{ground truth clustered outcome matrix} 
$\bfH = (h_{ij})_{i \in \cU^\star, j \in \cV^\star}$, where $
    h_{ij} = f(\bu_i, \bv_j), \quad i \in \cU^\star, j \in \cV^\star.$
We again emphasize that the rows and columns of $\bfH$ are indexed by central users $\cU^\star$ and central items $\cV^\star$ for simplicity. Using data, we form the \emph{unfilled clustered outcome matrix} $\overline{\bfH} \in \Reals^{|\cU^\star| \times |\cV^\star|}$, where for all $(i,j) \in \cU^\star\times \cV^\star$,
\begin{equation}\label{eq:def-bar-H}
    \bar h_{ij} = \begin{cases}
    \frac{1}{|\Omega_{ij}|} \sum_{(k, l) \in \Omega_{ij}} y_{kl} &\quad \text{if $|\Omega_{ij}| \geq N_n$;} \\
    \star &\quad \text{otherwise.}
    \end{cases}
\end{equation}
$N_n$ is a hyperparameter that governs whether there are ``enough'' observations available to produce a sufficiently accurate estimate of $h_{ij}$ via simple averaging. The choice of this hyperparameter will be specified later. 

\paragraph{Imputation.} We will impute the missing entries of $\overline{\bfH}$ using the available ones. The resulting matrix of this imputation, also referred to as \emph{imputed clustered outcome matrix}, will be denoted by $\widehat{\bfH}$. 
Let $i,j \in [n]$, if $\bar{h}_{ij} \neq \star$, then we simply set $\hat{h}_{ij} = \bar h_{ij}$. If $\bar{h}_{ij} = \star$, then we impute the missing entry by proceeding as follows:
\begin{itemize}
    \item[\emph{(i)}] First, we search over the sets $\cU^\star$ and $\cV^\star$ to find subsets of $d$ distinct central users $\cI$ and $d$ distinct central items $\cJ$ such that $\sigma_d(\overline{\bfH}_{\cI, \cJ})$ is maximized and 
\begin{equation} \label{eq:sep}
      \forall (i',j') \in \cI \times \cJ,  \ \ \bar h_{i, j'} \neq \star, \bar h_{i', j'} \neq \star, \bar h_{i',j} \neq \star. 
\end{equation}
    \item[\emph{(ii)}] Then, we define
\begin{equation}\label{eq:impute}
    \hat h_{ij}  = \begin{cases}
    \overline{\bfH}_{i, \cJ}\left(\overline {\bfH}_{\cI,  \cJ}\right)^{\dagger} \overline{\bfH}_{\cI, j}  & \text{if }\cI, \cJ \text{ are found};\\
    \star &  \text{otherwise.}
    \end{cases} 
\end{equation}
\end{itemize}
As we shall show later, $(\cI, \cJ)$ are guaranteed to be found, thus ensuring the correctness of the imputation procedure.

\begin{figure}
    \centering
    \includegraphics[width=1\linewidth]{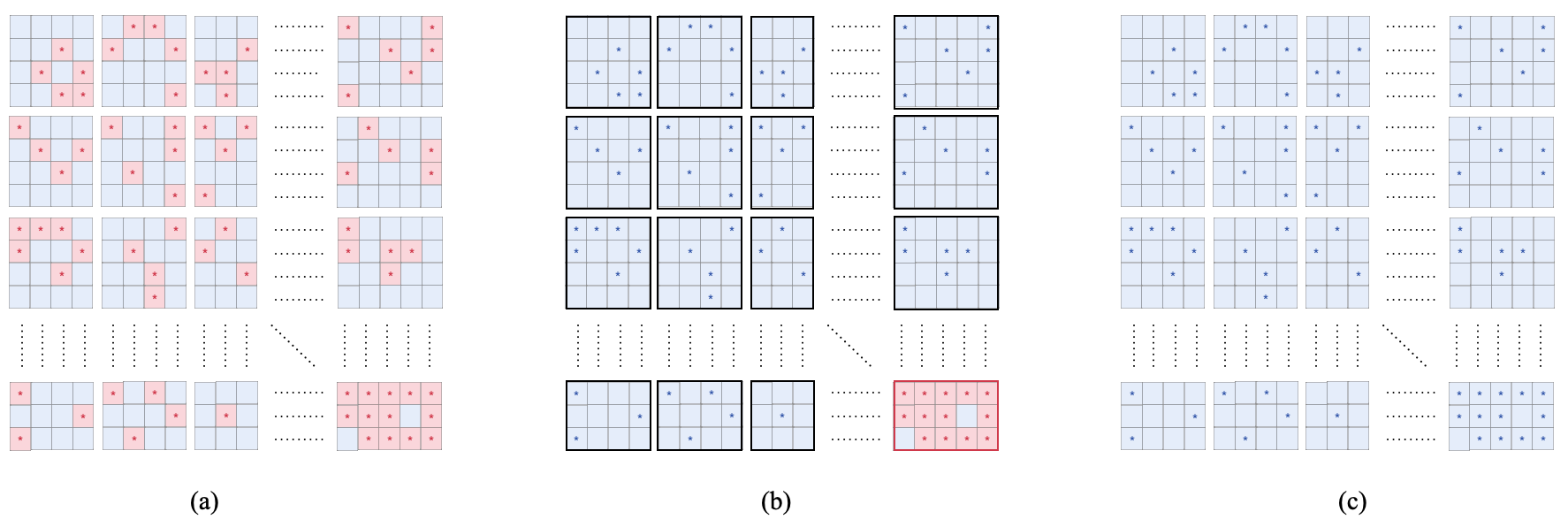}
    \caption{\emph{Visual depiction of the second stage of MNN.} The blue cells correspond to observed or imputed entries, and the red cells correspond to unobserved entries. The matrix illustrated in (a) corresponds to the outcomes matrix after rearranging its rows and columns. The matrix in (b) corresponds to the resulting matrix after the denoising step using  \eqref{eq:def-bar-H}. The matrix illustrated in (c) corresponds to the filled entries after using our the imputation procedure using \eqref{eq:impute}.}
    \label{fig:enter-label}
\end{figure}

\paragraph{Outcome Prediction.} Once $\widehat{\bfH}$ is constructed, predicting outcomes is simple: For any $i, j \in [n]$,  there exists $k \in \cU^\star$ and $\ell \in \cV^\star$ such that $i \in \cU_k$ and $ j\in \cV_\ell$, we then set $\hat{x}_{ij} = \hat{h}_{kl}$. This completes the description of $\MNN$.

\section{Prediction Guarantee of MNN}

Our main result is Theorem \ref{thm:prediction} which corresponds to the finite-sample performance guarantee of $\MNN$ in terms of its prediction error at the \emph{user-item} level. To establish this result, our analysis relies on three key ingredients: First, we show in that the joint clustering $\&$ denoising procedure yields good estimates for the outcomes of the central user-item pairs whenever they are revealed (see Proposition \ref{prop:clustering-denoising}); second, we characterize the pattern of the revealed central user-item pairs (see Proposition \ref{prop:imputation-1}); third, we show that the imputed central user-item pairs are accurate (Proposition \ref{prop:imputation2}). 



\subsection{Clustering, Denoising.} We denote the set of revealed user-item clusters by
$
    \Omega^\star = \lbrace (i,j): i \in \cU^\star, j \in \cV^\star, \quad \vert \Omega_{ij}\vert  \ge   N_n \rbrace. 
$
In Proposition \ref{prop:clustering-denoising}, we make the choice of $N_n$ precise and provide a finite-sample  guarantee on the error rate between the revealed entries of $\overline{\bfH}$ and $\bfH$.

\begin{proposition}\label{prop:clustering-denoising} Let $\delta \in (0,1)$. Suppose we choose $N_n \gtrsim  n^2 \rho_n \varepsilon_n^{2d-2}$
then the event: $\forall (i,j) \in \Omega^\star,$
\begin{equation}
     \vert \bar h_{ij} -  h_{ij} \vert \lesssim L B \varepsilon_n + \sqrt{\frac{\sigma^2}{n^2\rho_n} \! \left( \frac{2}{\varepsilon_n}\right)^{2d-2}\! \log\left(\frac{e \, n}{\delta}\right)} ,
\end{equation}
holds with probability at least $1- \delta$, provided $n \gtrsim d \log\left( e\,n /\delta \right)$ and 
\begin{equation}
            \varepsilon_n \gtrsim \frac{1}{\rho_n^2}\sqrt{\frac{d^3}{n} \log\left( \frac{e \, n}{\delta}\right)}.\label{condition:eps-1}
\end{equation}
\end{proposition}

We prove Proposition \ref{prop:clustering-denoising} in Appendix \ref{sec:proof-prop-clustering-denoising}. 


\subsection{Observations Pattern of $\overline{\bfH}$.} 
In order to ensure correctness of the procedure by which we construct  $\widehat{\bfH}$, the mappings $\theta(\cdot)$ and $\phi(\cdot)$ must be ``non-degenerate'' in some appropriate sense so that the search for $\cI$ and $\cJ$ is successful, and that the smallest singular value of $\overline{\bfH}_{\cI, \cJ}$  is bounded away from $0$ independently from $n$. This will in turn ensure that observation bias may still be exploited to impute the missing entries. Below, we precisely state what we mean by ``non-degenerate''.

\begin{definition}[Non-degeneracy]\label{def:non-degen}
    We say that a mapping $\theta(\cdot)$ from $\cS^{d-1}$ to $\RR^d$ is non-degenerate with parameters $m,\kappa > 0$ if it satisfies the following two properties: \emph{(a)} for all $\bx \in \cS^{d-1}$, $\Vert \theta(\bx)\Vert_2 \ge m \Vert \bx \Vert_2$; \emph{(b)}  for all $\bx, \by \in \cS^{d-1}$, $\Vert \tilde\theta(\bx) - \tilde\theta(\by) \Vert_2  \ge \kappa \Vert \bx - \by \Vert_2$, where we define $\tilde{\theta}({\bx}) = \norm{\theta(\bx)}^{-1}\theta(\bx)$.
\end{definition}

Now, we are ready to present Proposition \ref{prop:imputation-1}, in which we describe the pattern of revealed entries in $\overline{\bfH}$, provided that $\theta(\cdot)$ and $\phi(\cdot)$ are ``non-degenerate''. This in turn ensures that $\Impute(\overline{\bfH}, i, j)$ will successfully terminate with high probability. 

\begin{proposition}\label{prop:imputation-1} Assume that $\theta(\cdot)$ and $\phi(\cdot)$ are non-degenerate with parameters $m,\kappa > 0$, and that $N_n$ is chosen as in Proposition \ref{prop:clustering-denoising}. Let $i \in \cU^\star, j \in \cV^\star$ be such that $\bar{h}_{i,j} = \star$, (i.e., $(i,j)$ is a non-revealed user-item cluster in $\overline{\bfH}$). Then, for all $\delta \in (0,1)$, the event: there exists $\cI \subset \cU^\star, \cJ \subseteq \cV^\star$ with $\vert \cI \vert = \vert \cJ\vert = d$, such that $\sigma_d(\bfH_{\cI, \cJ}) \gtrsim m^2 \kappa^4,$ and 
\begin{align}
     \!\!\forall (i',j') \in \cI \times \cJ,\quad  \bar{h}_{ij'} \neq \star, \ \ \bar{h}_{i'j'} \neq \star, \ \ \bar{h}_{i'j} \neq \star,
\end{align}
holds with probability at least $1- \delta$, provided that $\varepsilon_n$ satisfies 
\begin{align}
     \left( \frac{\varepsilon_n}{2}\right)^{d-1} & \gtrsim \frac{1}{n}\left (\sqrt{\frac{1}{\rho_n}\log\left( \frac{e \, n}{\delta} \right) } + \log\left( \frac{e \, n}{\delta}\right) \right), \nonumber \\
    \varepsilon_n  & \lesssim  \left(\frac{\kappa^4 (m^2 \wedge 1)}{L(B \vee 1)} \wedge 1 \right). \label{condition:eps-2}
\end{align}
\end{proposition}

The proof of Proposition \ref{prop:imputation-1} is given in Appendix \ref{app:imputation-1}. It is worth noting that the ``non-degeneracy'' property allows us to rule out pathological cases for which it may not be possible to impute the missing entries from the observed ones, at least using our approach. The need for such property is apparent in the proof, and we refer the reader to Appendix \ref{app:imputation-1} to see how this property comes into our analysis. 

\subsection{The completed matrix $\widehat{\bfH}$} Next, we present a finite-sample error bound on the imputation error of all entries including the missing ones in $\overline{\bfH}$. 
\begin{proposition}\label{prop:imputation2} Let $\delta \in (0,1)$. Under the setup of Proposition \ref{prop:imputation-1}, the event: $\forall (i,j) \in \cU^\star \times \cV^\star,$
\begin{align}
         \vert \hat{h}_{ij} - h_{ij} \vert \lesssim  L B \varepsilon_n  + \sqrt{\!\frac{\sigma^2}{n^2\rho_n} \!\!\left( \frac{2}{\varepsilon_n}\right)^{2d-2}\!\!\!\! \log\left(\frac{e \, n}{\delta}\right)} 
\end{align}
holds with probability at least $1- \delta$, 
provided that $n \gtrsim d \log(e \, n /\delta)$, and  $\varepsilon_n$ satisfies  
\eqref{condition:eps-2}, and 
    \begin{align}
               \left(\frac{\varepsilon_n}{2} \right)^{d-1} &  \!\!\gtrsim  \frac{d}{m^2 \kappa^4 n} \! \left ( \!\!\sqrt{\frac{1}{\rho_n}\log\left( \frac{e \, n}{\delta} \right) } + \log\left( \frac{e \, n}{\delta}\right)\! \! \right), \nonumber \\   \varepsilon_n & \lesssim \frac{\kappa^4 (m^2 \wedge 1)}{d L(B\vee 1)} \wedge 1 
    \end{align}
\end{proposition}

The proof of Proposition \ref{prop:imputation2} is based on Proposition \ref{prop:clustering-denoising} and Proposition \ref{prop:imputation-1}. We give its proof in Appendix \ref{app:imputation-2}.

\subsection{Main prediction error guarantee} Finally, we are ready to present the main performance guarantee satisfied by $\MNN$. To simplify the exposition of our result, we will consider $\rho_n = n^{-\beta}$ for some $\beta>0$. 
\begin{theorem}\label{thm:prediction} Let $\delta \in (0,1)$. Under the setup of Proposition \ref{prop:imputation-1}, with choice the $\varepsilon_n = 2 n^{-(2 - \beta)/(d-1)}$, the event: 
\begin{align}\label{eq:thm:error-rate}
     \max_{i,j\in [n]}\vert \hat{x}_{ij} - x_{ij}\vert \le K \,  n^{-\frac{2-\beta}{2d}}\sqrt{\log\left(\frac{e\, n}{\delta}\right)},  
\end{align}
holds with probability at least $1-\delta$, provided that $d> 9/4$, $\beta< 1/4$, and 
\begin{align}\label{eq:thm:condition}
    n \ge \left( \frac{C  L B d^{3} }{m^2 \kappa^4}    \log\left( \frac{e \, n}{\delta}\right) \right)^{2d}.
\end{align}
Here, constant $K = \textup{poly}(d,L,B, \sigma, m^{-1}, \kappa^{-1})$, and $C>0$ is a universal constant.
\end{theorem}

 Observe that condition on $n$ in Theorem \ref{thm:prediction} can be verified\footnote{We can verify by using the Lambert function $W_0$, which satisfies $W_0(x) = y$ with $x = y^y$ for $x \ge 0$. Additionally, we have $W_0(x) \ge \log(x) - \log\log(x) + \frac{\log\log(x)}{\log(x)}$, for all $x \ge e$. (see \cite{hoorfar2008inequalities}).} if  $d = O(\log(n) - \log(\log (n)))$ 
where $O(\cdot)$ may hide polynomial dependencies on $L, B, m^{-1}$ and $\kappa^{-1}$. We shall now present the proof of Theorem \ref{thm:prediction} which is a direct consequence of Proposition \ref{prop:imputation2} and Lemma \ref{lemma:distance}.

\paragraph{Proof of Theorem \ref{thm:prediction}.} 

Fix $i,j \in [n]$. We know, thanks to our clustering procedure, that there exist clusters $k \in \cU^\star$, $ \ell \in \cV^\star$ such that $i \in \cU_k$ and $j \in \cV_\ell$. Thus, we may decompose the prediction error as follows 
\begin{align}
    x_{ij} - \hat{x}_{ij}  =   x_{ij} -  h_{k\ell}  +   h_{k\ell} - \hat{h}_{k\ell}. 
\end{align}

\medskip 

\underline{\emph{Step 1. (Bounding the first term).}} First, we note that 
\begin{align}
    \vert h_{k\ell} - x_{ij} \vert & = \vert f(\bu_k, \bv_\ell) - f(\bu_i, \bv_j) \vert  \\
    & \le \sqrt{d}BL (\Vert \bu_k - \bu_i \Vert_2 + \Vert \bv_\ell - \bv_j \Vert_2). 
\end{align}
We know, under our clustering procedure, that $\hat{d}_{ki}^\bu \le 6 \varepsilon_n$ and $\hat{d}_{\ell j}^\bv \le 6 \varepsilon_n$. Hence, using Lemma \ref{lemma:distance}, we obtain that the event: 
\begin{align}
     \vert h_{k\ell} - x_{ij} \vert & \le \sqrt{d}BL(\Vert \bu_k - \bu_i \Vert_2 + \Vert \bv_\ell - \bv_j \Vert_2 ) \\
     & \le 14\sqrt{d}BL\varepsilon_n 
\end{align}
holds with probability $1-\delta/(2n^2)$, provided that 
provided that 
\begin{align}
    \varepsilon_n & \ge \frac{c}{\rho_n^2}\sqrt{\frac{d^3}{n} \log\left( \frac{e\,n}{\delta} \right)} = c n^{2\beta - \frac{1}{2}} \sqrt{d^3\log\left( \frac{e\,n}{\delta} \right)} \label{eq:c1}\\ 
    n & \ge C d\log\left(\frac{e \, n }{\delta}\right), \label{eq:c2}
\end{align}
where $c, C >0$ are universal constants, and recall that $\rho_n = n^{-\beta}$. Note that if we wish to obtain a non-trivial error rate, then condition \eqref{eq:c1} implies that we require
$
    \beta < \frac{1}{4}.
$

\medskip 

\underline{\emph{Step 2. (Bounding the second term).}} Second, we invoke Proposition \ref{prop:imputation2}, to obtain that the event: 
\begin{align*}
    & \vert \hat{h}_{k\ell} - h_{k\ell} \vert \le c \left(\frac{d^3B^2}{(m^2\kappa^4 \wedge m^4\kappa^8)} \wedge 1 \right)   \left( L B \varepsilon_n  + \frac{\sigma}{n} \left(\frac{2}{\varepsilon_n}\right)^{d-1} \sqrt{ \frac{1}{\rho_n}\log\left(\frac{e\, n}{\delta}\right)}  \right)
\end{align*}
holds with probability at least $1 - \delta/(2n^2)$, provided that  \eqref{eq:c1}, \eqref{eq:c2}, and   
\begin{align}
    \left( \frac{\varepsilon_n}{2} \right)^{d-1} & \ge \frac{Cd}{m^2 \kappa^4 n}\left (\sqrt{\frac{1}{\rho_n}\log\left( \frac{e \, n}{\delta} \right) } + \log\left( \frac{e \, n}{\delta}\right) \right)  \nonumber \\
    & = \frac{Cd}{m^2 \kappa^4 n}\left( n^{\beta/2} \log^{1/2}\left( \frac{e n}{\delta}\right) + \log\left( \frac{e n}{\delta}\right)\right)  \label{eq:c3}\\
    \varepsilon_n & \le C \left( \frac{\kappa^4 (m^2 \wedge 1)}{d L(B\vee 1)} \wedge 1 \right) \label{eq:c4}
\end{align}
where $C>0$ is some universal constant.

\medskip 

\underline{\emph{Step 3. (Putting everything together).}} To complete our analysis, we choose $\varepsilon_n = 2n^{-\alpha}$, and find the optimal choice of $\alpha$ that minimizes prediction error. Under the two events described above in \emph{step 1} and \emph{step 2}, ignoring the problem dependent constants, it is clear that the prediction error scales as follows with $n$, 
\begin{align}
    \vert \hat{x}_{ij} - x_{ij} \vert & \lesssim \varepsilon_n + \frac{1}{n} \left( \frac{2}{\varepsilon_n}\right)^{d-1} \sqrt{\frac{1}{\rho_n} \log\left(\frac{e \, n}{\delta}\right)}  \\
    & \lesssim \frac{1}{n^{\alpha}} + n^{\alpha(d-1) - 1 + \frac{\beta}{2}} \sqrt{ \log\left(\frac{e \, n}{\delta}\right)}.
\end{align}
Clearly, choosing 
$
    \alpha = \frac{2 - \beta}{2d},
$
yields the minimal error rate 
\begin{align}
    \vert \hat{x}_{ij} - x_{ij} \vert  \lesssim n^{- \frac{2 - \beta}{2d}}\sqrt{\log\left( \frac{e\,n}{\delta}\right)}.
\end{align}
However, for such choice of $\alpha$ to be valid, we need to verify that the conditions \eqref{eq:c1}, \eqref{eq:c3}, and \eqref{eq:c4} are verified. Now, we can easily verify that condition \eqref{eq:c1} holds if 
\begin{align}
    \beta < \frac{d-1}{4d - 1} \left( 1 - \frac{d}{d-1} \frac{\log\left( c^2 d^3 \log\left(e n/\delta\right)\right)}{\log(n)}  \right)
\end{align}
Next, we can also easily verify that condition \eqref{eq:c3} holds if 
\begin{align}
    \beta &\le 2   \left(1  - \frac{\log\left(\frac{C d}{2 m^2\kappa^4} \sqrt{\log\left( \frac{e n }{\delta}\right)}\right)}{\log(n)}\right) \\
    \frac{1}{d} & \ge  \frac{\log(C d/(2 m^2\kappa^4)) + \log\log(en/\delta)}{\log(n)} 
\end{align}
Finally, we easily verify that condition \eqref{eq:c4} holds if 
\begin{align}
    \beta & \le  2 \left( 1 +  \frac{d \log\left(\frac{C}{2} \left( \frac{\kappa^4 (m^2 \wedge 1)}{d L(B\vee 1)} \wedge 1 \right)\right)}{\log(n)}\right).  
\end{align}
In summary, we see that when 
\begin{align}
    n \ge \left( \frac{C' d^{3} L B }{m^2 \kappa^4}    \log\left( \frac{e \, n}{\delta}\right) \right)^{2d}, \quad d > 9/4, \quad \beta < \frac{1}{4}, 
\end{align}
then all the conditions above are satisfied, and thus  \eqref{eq:c1}, \eqref{eq:c3}, and \eqref{eq:c4} are verified.

\section{Comparison with lower bounds in supervised learning} \label{sec:param-rate}

\paragraph{Parametric rates in supervised learning.} Suppose we knew exactly the latent features $(\bu_i)_{i \in [n]}$,  $(\bv_j)_{j \in [n]}$, then predicting outcomes of interest from noisy observations amounts to solving a supervised learning task where we recall that $f: \RR^{2d} \mapsto \RR$ is a bounded Lipschitz function that is a priori unknown (we recall from section \ref{sec:problem} that we adopt a model of the form $y_{ij} = f(\bu_i, \bv_j) + \xi_{ij}$). In this classical supervised learning setting, we know, for example from Theorem 1 in \cite{stone1982optimal}, that the optimal achievable parametric prediction error rate is 
\begin{align}
    \max_{i,j \in [n]}\vert \hat{x}_{ij} - x_{ij} \vert = \widetilde{\Theta}\left( N^{-\frac{1}{2 + 2d}}\right)
\end{align}
where $N$ would correspond to the number of observations and $\widetilde{O}(\cdot)$ may hide logarithmic dependencies on $N$. Let us further add that the $k$-nearest neighbor is an example of supervised learning method that achieves such rate, for example see \cite{chenshah}.

\paragraph{Comparison with parametric rates in supervised learning.} The latent vectors $(\bu_i)_{i \in [n]}$,  $(\bv_j)_{j \in [n]}$ in our proposed model are not observed as they represent confounding factors. Despite this additional challenge, the proposed algorithm $\MNN$ still attains a competitive rate in comparison with the parametric one for the supervised learning setting described above. Indeed, we provide the following result: 

\begin{lemma}\label{lem:prediction-param-rate}
Consider the setup of Theorem \ref{thm:prediction}. Let $N$ denote the expected number of observed noisy entries of $\bfX$. 
Then for all $\delta \in (0,1)$, the event
\begin{align}
    \max_{i,j \in [n]}\vert \hat{x}_{ij} - x_{ij}\vert =  \widetilde{O}\left(N^{-\frac{1}{2d}} \sqrt{\log\left(\frac{e}{\delta}\right)}\right)
\end{align}
holds with probability at least $1-\delta$, provided conditions of Theorem \ref{thm:prediction} hold. Here 
$\widetilde{O}(\cdot)$ hides logarithmic dependencies on $N$ and polynomial dependencies in 
$d,L,B, \sigma, m^{-1},$ and $ \kappa^{-1}$.
\end{lemma}
We omit the proof of Lemma \ref{lem:prediction-param-rate} as the result follows immediately from Theorem \ref{thm:prediction}. To see that, note that under the proposed model, according to \eqref{eq:prob-obs}, the expected number of observation $N \approx n^{2-\beta}$. Rewriting this as $n \approx N^{-1/(2 - \beta)}$ and plugging it in the error rate expression \eqref{eq:thm:error-rate} 
in Theorem \ref{thm:prediction} yields the desired result.   

\medskip 

In conclusion, estimating $x_{ij}$ using $\MNN$ achieves a prediction error rate that is competitive with the parametric rate for sufficiently slowly decaying $\rho_n$, despite the following unconventional challenges: (i) the algorithm does not have direct access to the latent features $(\bu_i)_{i \in [n]}$ and $(\bv_i)_{i \in [n]}$, but instead has to estimate them; (ii) samples are drawn in a biased manner as opposed to the uniform sampling of datapoints over the domain, which is typically assumed.

\medskip

It is worth nothing that Theorem \ref{thm:prediction} assumes that $f$ factorizes as in \eqref{eq:rep-outcome} with number of factors being $d$. Indeed as $d$ scales it becomes a generic Lipschitz function but then the dependence on $d$ in Theorem \ref{thm:prediction} would start dominating. The assumption \eqref{eq:rep-outcome} is also the reason why $N^{-\frac{1}{2d}} \ll N^{-\frac{1}{2+2d}}$ is not a contradiction.

\section{Experiments}\label{sec:experiments}

\subsection{Real-world data}

In this section, we provide few experimental results on a real-world dataset to illustrate the performance gains promised by our theoretical findings. 

\medskip 

\textbf{Data.} We used a dataset obtained from Glance\footnote{Glance (www.glance.com) is a smart lock screen that aims to personalize the screen.} detailing the interaction of users with different pieces of content, considering a partially observed matrix where rows represent users, columns represent the different pieces digital content, and outcomes are the natural logarithm of the duration (in seconds) that the user interacted with the content. The dataset we considered had 1014696 observed measurements for 1305 users and 1471 content items. The sparsity of the dataset was thus $53\%$ and consistent with the observation model described in our setup. 

\medskip 

\textbf{Algorithms compared.} We utilize a practical implementation of the proposed $\MNN$ algorithm where (i) $\bfA$ is used in place of $\tilde\bfA$ in the distance estimation step and (ii) the matrix completion from the clustered outcome matrix is done using SVD with known rank. Using such a practical implementation of the algorithm acts a proof of concept -- letting us determine the added value of the extra information gained from the two stage approach. The details of such implementation are provided in Section \ref{sec:exp-mnn-implementation}. We compare the performance of such implementation against a modified version of $\USVT$, namely a version that has knowledge of the rank. We sought to further compare against $\SNN$ of \cite{agarwal2021causal}, an algorithm achieving state-of-the-art results. However, due to prohibitive computational requirement, $\SNN$ is impractical for values of $n$ as large as in the Glance dataset. Instead, we compare $\SNN$ with $\MNN$ on smaller size synthetic data as discussed in Section \ref{sec:comparison-snn}.

\medskip 

\textbf{Metrics.} 
We use $R^2$ score, Mean Squared Error (MSE), Mean Absolute Error (MAE), and Maximum Across All Entries (max error) to assess performance of estimated outcome matrices. We directly compare the distribution of estimated entries to the distribution of true outcomes to assess the 
bias of the algorithms. 

\medskip 

\textbf{Experiment setup.} We split the data into $90/10$ train/test sets at random and repeat the experiment 10 times. We determined the best estimate of the rank of the true outcome matrix and the rank of the observation pattern using 9-fold cross-validation with $\MNN$. We used 16-fold cross-validation to separately tune the other hyperparameters. Next, we ran both $\MNN$ and modified $\USVT$ on the training set in each experimental run. Using the tuned hyperparameters, we ran both algorithms for 10 experimental runs using different train/test splits to calculate performance metrics.

\medskip 

\begin{figure}[ht!]
    \centering
    \includegraphics[width=0.9\columnwidth]{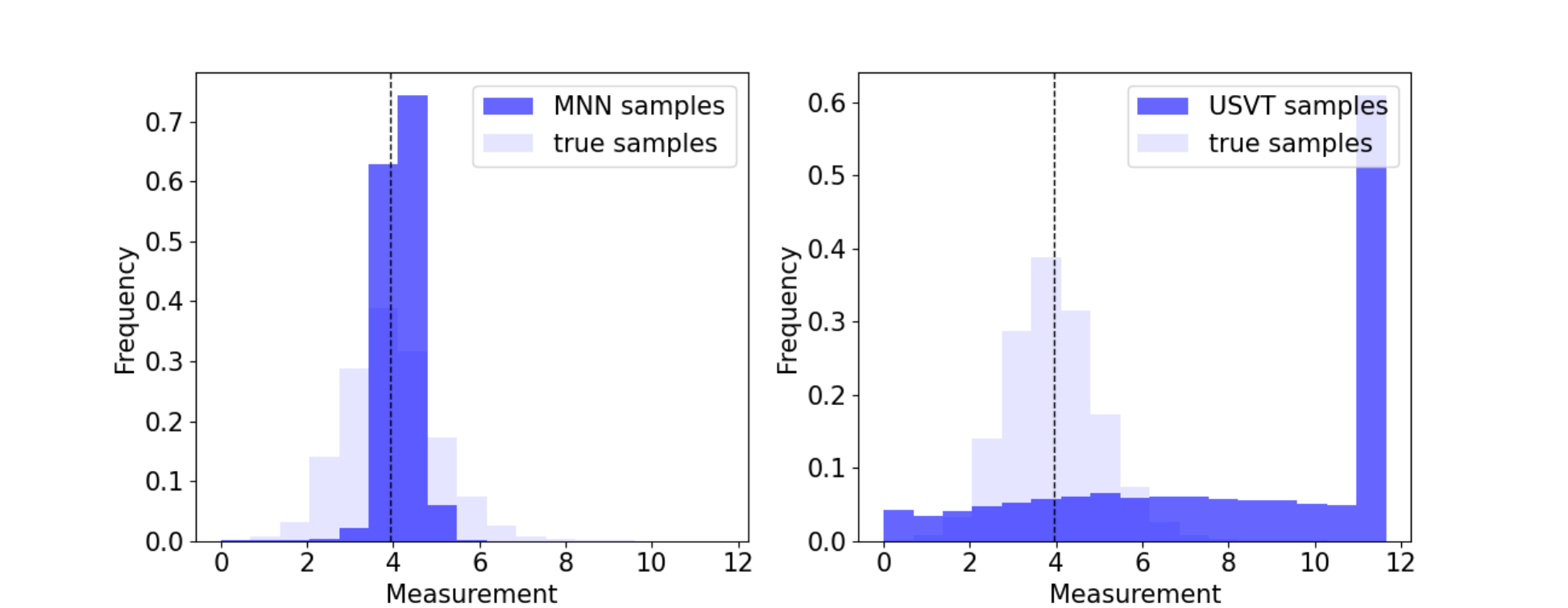}
    \caption{Distributions of test set estimates from $\MNN$ and modified $\USVT$ for real-world data. The dashed line represents the mean of the true distribution. As can be seen $\MNN$ is reasonably accurate while $\USVT$ is large bias.}
    \label{fig:real_hist}
\end{figure}

\begin{figure}[ht!]
    \centering
    \includegraphics[width=.9\columnwidth]{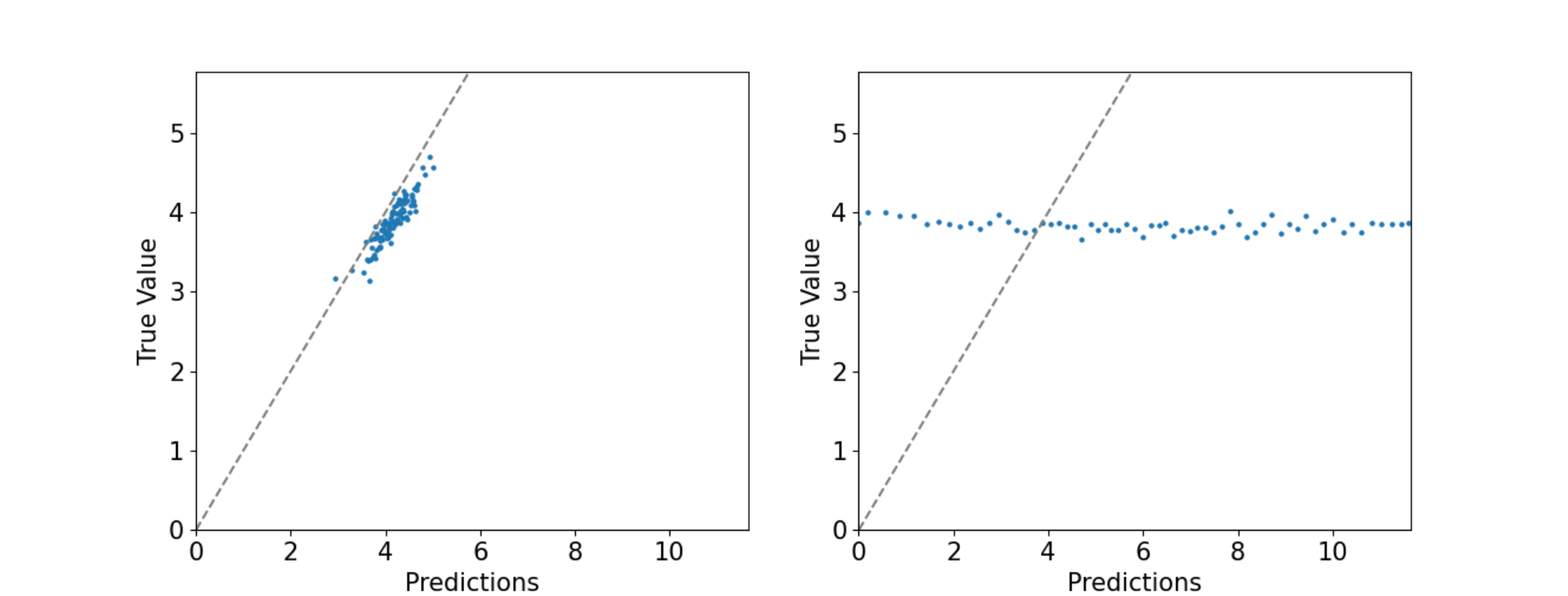}
    \caption{Predictions vs true outcomes for real-world data using $\MNN$ (left) and modified $\USVT$ (right). The dashed line represents perfect prediction. As can be seen $\MNN$ is reasonably accurate while $\USVT$ is far from the ground truth.}
    \label{fig:real_bias}
\end{figure}

\textbf{Results.} As we can see from Fig. \ref{fig:real_bias}, the estimates from modified $\USVT$ are extremely biased. The estimates from $\MNN$, however, appear to be minimally biased and inline with ground truth. Moreover, from Fig. \ref{fig:real_hist}, we can see that the estimates made by modified $\USVT$ are very sensitive to outliers in the data, while the estimates from $\MNN$ are not. We also observe from Table \ref{table:real_performance} that the $R^2$, MSE, MAE, and max error of $\MNN$ are far better than that of modified $\USVT$. Specifically the MSE of $\MNN$ is > 28x better compared to modified $\USVT$. That is, $\MNN$ works significantly better on MNAR data.

\begin{small}
    \begin{sc}
        \begin{table}[ht!]
            \begin{center}
                \renewcommand{\arraystretch}{1.25}
                    \begin{tabular}{ ccccc } 
                         \toprule
                         Algorithm & $R^2$ & MSE & MAE & Max Error\\ 
                         \toprule
                         \MNN & $\mathbf{.039 \pm .038}$ & $\mathbf{1.15 \pm .054}$ & $\mathbf{.843 \pm .021}$ & $\mathbf{6.49 \pm .405}$\\
                         \midrule
                         \USVT & $-26.0 \pm .435$ & $32.4 \pm .392$ & $4.95 \pm .032$ & $11.6 \pm .053$\\
                         \bottomrule
                    \end{tabular}
            \end{center}
            \caption{Comparison of performance of $\MNN$ and USVT on Glance data. As can be seen, MSE for $\MNN$ is >28x better.}
            \label{table:real_performance}
        \end{table}
    \end{sc}
\end{small}

\subsection{Synthetic Data} \label{sec:exp-synthetic}

\textbf{Data.} We performed several experiments with different numbers of datapoints $n$ and sparsity values $\rho_n$, all of which had data generated using the following steps. First, we set $d = 5$ and generated $d$-dimensional latent factors from the unit sphere  for each of the $n$ users and items. Next, we generated the matrix $\bfP$ of probabilities and observation matrix $\bfA$ as described in our model. We defined the outcome matrix $\bfX = \theta(\bfU)\phi(\bfV)^\top$ where $\theta(\bfU) = \exp({\sqrt{d}\bfU})
$
and $
    \phi(\bfV) = \exp({\sqrt{d}\bfV}).
$ and $\exp(\cdot)$ is applied entry-wise.
We then generated the subgaussian noise terms $\xi_{ij}$ by sampling $n^2$ entries i.i.d from a standard normal distribution. Finally, the partially-observed noisy outcome matrix $\bfY$ was generated by setting $y_{ij} = x_{ij} + \xi_{ij}$ if $a_{ij} = 1$ and $y_{ij}$ to be a null value otherwise.

\medskip 

\textbf{Algorithms compared.} We compare the performance of the proposed algorithm against a standard matrix completion algorithm $\USVT$. In the experiments, we utilize a practical implementation of the proposed $\MNN$ as explained in Section \ref{sec:exp-mnn-implementation}. We compare against a modified version of $\USVT$ where knowledge of the rank of the matrix is known, allowing for more comparable performance against $\MNN$ which utilizes the matrix rank. We note that $\MNN$ is designed to handle biased data while the modified $\USVT$ algorithm is not. 

\medskip 

\textbf{Metrics.} We use $R^2$ score, MSE, MAE, and max error to assess the similarity of the estimated full outcome matrix with the true outcome matrix. We directly compare the distribution of estimated entries to the distribution of true outcomes to assess the bias of the algorithms. 

\medskip 

\textbf{Experiment setup.} We performed two different experiments with the described algorithms, using 16-fold cross-validation to tune the hyperparameters. In the first experiment, we tried 3 values of $n \in \{100, 562, 3162\}$. We set $\rho_n = n^{-1/4}$. We then ran both $\MNN$ and modified $\USVT$ with these values of $n$ and compared the resulting estimates. In the second experiment, we set $n = 1000$ and tested 3 values of $\beta \in \{0, 1/6, 1/4\}$. We set $\rho_n = n^{-\beta}$. We then again ran both $\MNN$ and modified $\USVT$ and compared the bias of the resulting estimates. We concluded by running 10 experimental repeats of $\MNN$ and modified $\USVT$ using data generated with different random seeds for each of these $\rho_n = n^{-\beta}$ values. These final runs were used to calculate the $R^2$, MSE, MAE, and max error as a function of $n$ for each $\beta$. 

\medskip 

\begin{figure}[ht]
    \centering
    \includegraphics[width=\columnwidth]{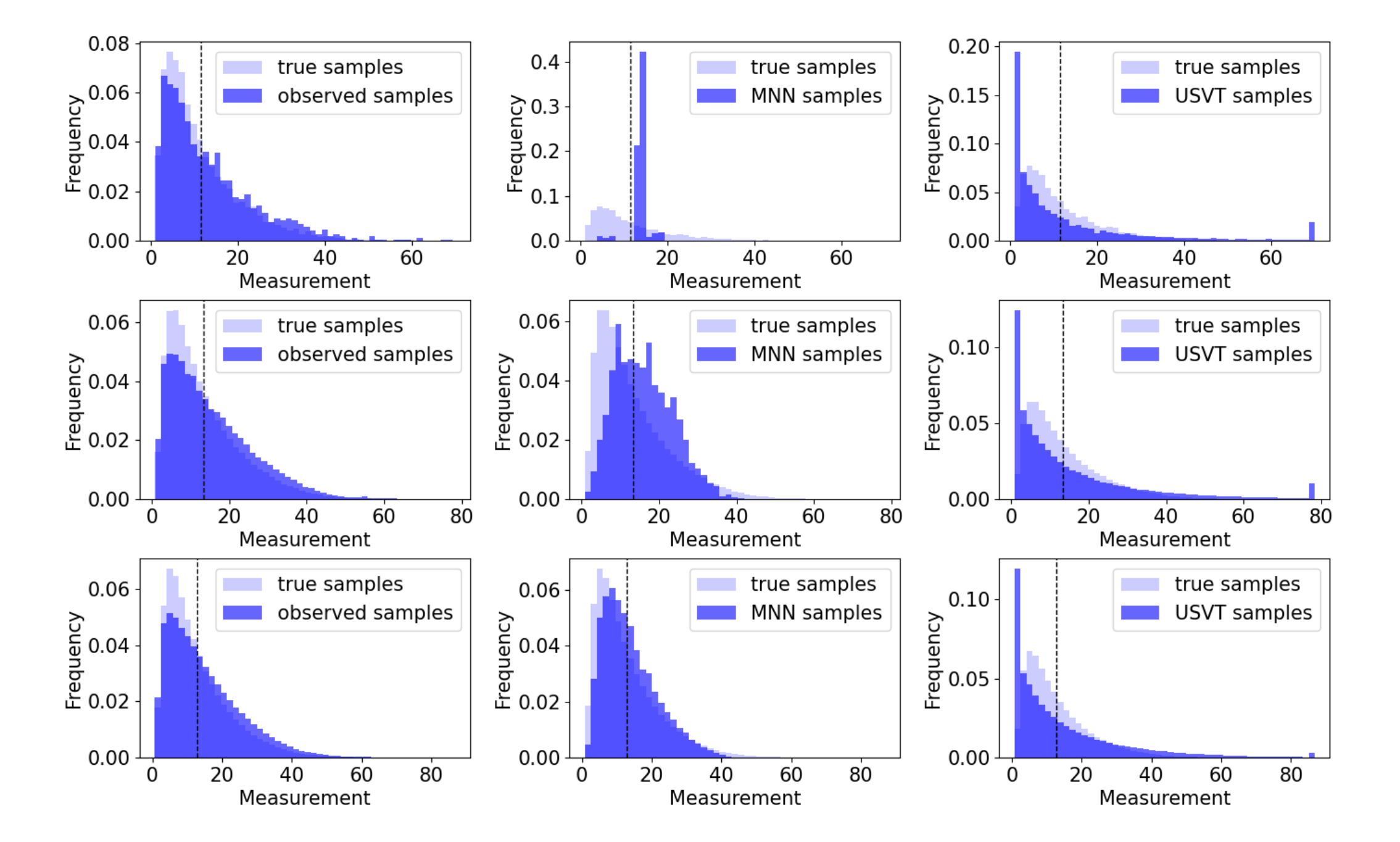}
    \caption{Distributions for synthetic data generated with $n = 100, 562$, and $3162$ users and items (top to bottom). The dashed line represents the mean of the true distribution.}
    \label{fig:synthetic_hist}
\end{figure}

\begin{figure}[ht]
    \centering
    \includegraphics[width=\columnwidth]{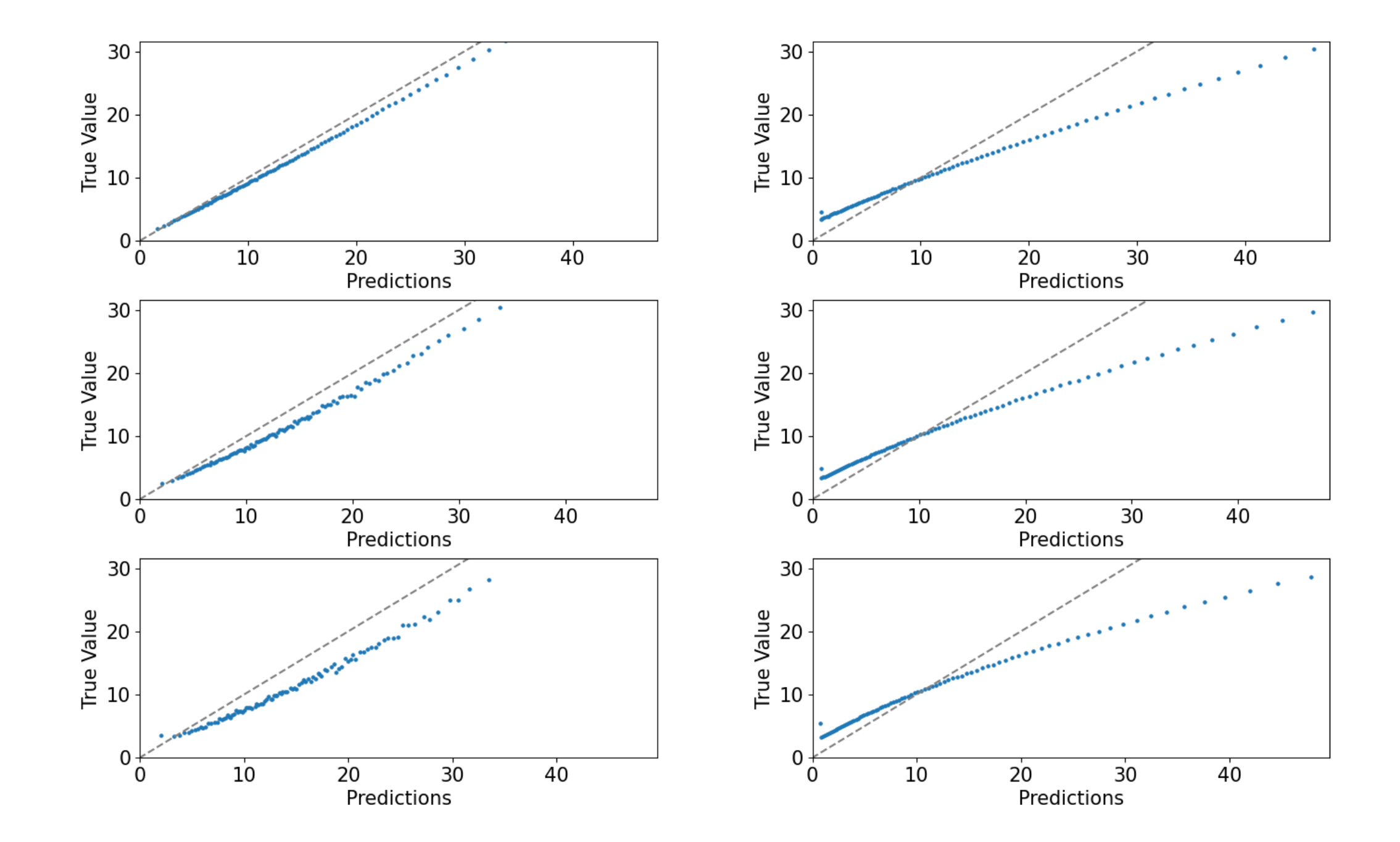}
    \caption{Predictions compared to true values for $\MNN$ (left) and modified $\USVT$ (right) with $\beta = 0, \beta = 1/6$, and $\beta = 1/4$ (top to bottom). The dashed line represents perfect prediction.}
    \label{fig:synthetic_bias}
\end{figure}

\textbf{Results.} Before comparing the performance of $\MNN$ and modified $\USVT$ on the synthetic dataset, we examine the bias of the estimates for the full outcome matrix in both cases (experiments 1 and 2). As can be seen in Fig. \ref{fig:synthetic_hist},the distribution of estimates generated by $\MNN$ better approximates the true distribution of outcomes as the number of datapoints increases. Moreover, as $n$ increases, the shape of the distribution indicates that the results are less biased. However, the estimates generated by modified $\USVT$ do not display this same trend and do not appear to become less biased as the number of datapoints increases. This difference in bias can also be seen in Fig. \ref{fig:synthetic_bias} as $n$ is held constant at 1000. As $\rho_n$ decreases from 1 to $n^{-1/4}$, the bias of $\MNN$ barely changes but the bias of modified $\USVT$ increases significantly. 

\medskip 

Now, we consider $R^2$ score, MSE, MAE, and max error as metrics to compare the estimates made by $\MNN$ and modified $\USVT$ against the true outcomes. The results of this experiment can be seen in Table \ref{table:synthetic_performance}. We can see that across all these metrics, $\MNN$ outperforms modified $\USVT$. When comparing the $R^2$ scores, we see that the difference in the performance of the two algorithms remains relatively constant for different $\rho_n$ values. The results of these $R^2$-, MSE-, MAE-, and max error-based metrics all line up with the observations made about the relative bias of the estimates in the two algorithms, indicating that the latent factor clustering approach utilized by $\MNN$ is indeed de-biasing the data and leads to better estimates. 

\begin{small}
    \begin{sc}
        \begin{table}[ht!]
        \caption{$R^2$, MSE, MAE, and max error for matrix completion methods on synthetic datasets (average $\pm$ standard deviation across 10 experimental repeats).}
            \begin{center}   
                    \begin{tabular}{ ccccc } 
                         \toprule
                         & Algorithm & $\beta = 0$ & $\beta = 1/6$ & $\beta = 1/4$ \\ 
                         \toprule
                         \multirow{2}{4em}{$R^2$} & \MNN & $\mathbf{.831 \pm .003}$ & $\mathbf{.573 \pm .011}$ & $\mathbf{.394 \pm .032}$ \\\cline{2-5}
                         & \USVT & $.393 \pm .003$ & $.200 \pm .017$ & $-.015 \pm .014$ \\ 
                         \midrule
                         \multirow{2}{4em}{MSE} & \MNN & $\mathbf{14.1 \pm .291}$ & $\mathbf{35.6 \pm .900}$ & $\mathbf{50.5 \pm 2.67}$ \\\cline{2-5} 
                         & \USVT & $50.6 \pm .250$ & $66.7 \pm 1.42$ & $84.7 \pm 1.19$ \\ 
                         \midrule
                         \multirow{2}{4em}{MAE} & \MNN & $\mathbf{2.62 \pm .028}$ & $\mathbf{4.34 \pm .066}$ & $\mathbf{5.37 \pm .168}$ \\\cline{2-5} 
                         & \USVT & $4.61 \pm .008$ & $5.05 \pm .043$ & $5.57 \pm .037$ \\ 
                         \midrule
                         \multirow{2}{4em}{Max Error} & \MNN & $\mathbf{36.0 \pm 2.24}$ & $\mathbf{56.3 \pm 3.56}$ & $\mathbf{65.6 \pm 2.65}$ \\\cline{2-5} 
                         & \USVT & $42.0 \pm .616$ & $59.3 \pm 2.20$ & $66.5 \pm 1.44$ \\ 
                         \toprule
                    \end{tabular}
            \end{center}
        \label{table:synthetic_performance}
        \end{table}
    \end{sc}
\end{small}

\subsection{Comparison with $\SNN$}\label{sec:comparison-snn}

\paragraph{Experimental setup on the synthetic data.}  We compared our method to $\SNN$ \cite{agarwal2021causal}, a matrix completion algorithm for the MNAR setting. We only compared with $\SNN$ because it achieves state-of-the-art results. We compared the two algorithms on our synthetically generated data as described in section \ref{sec:exp-synthetic}. The implementation of $\MNN$ is as described in Section \ref{sec:exp-mnn-implementation}. We considered $\beta = 1/4$ and run the experiments with $n$ items and $n$ users where $n \in \lbrace 100, 200, 300, 1000 \rbrace$. 

\paragraph{Results.} The results (see Table \ref{table:exp-snn}) suggest that the performance of MNN is comparable to that of $\SNN$ for relatively small values of $n$, namely $n=100$ and $n=200$, with $\SNN$ slightly outperforming $\MNN$ on the MAE metric. However, we observed that as $n$ grows large the computational performance of $\SNN$ degrades rapidly in comparison with our implementation of $\MNN$. This is to be expected because $\SNN$ needs to solve a biclique search problem which is known to be a computationally hard problem.  

\medskip 

\textbf{Comparison with $\SNN$ on Real-data.} We tried to also compare with $\SNN$ on the Glance dataset. In this dataset, the number of users is $1305$ and the number of content items is $1471$. Unfortunately, $\SNN$ was not able to finish running in a reasonable time. Indeed, even on the synthetically generated dataset with only $300$ users and $300$ items, $\SNN$ could not finish running within 24 hours as reported in the table because it is not computationally efficient (see section 4.2. in \cite{agarwal2021causal}). To be able to compare the with our algorithm, we would require a distributed implementation of $\SNN$, which is beyond the scope of this work. This also means that our proposed algorithm is meaningful.

\begin{table}[ht!]
    \caption{MSE, MAE, and runtime for $\MNN$ and $\SNN$ on synthetic datasets (average $\pm$ standard deviation across 10 experimental repeats). $--$ means the method did not complete within $24$ hours. \vspace{.2cm}}
    \centering
    \renewcommand{\arraystretch}{1.25}
    \begin{tabular}{c ccccc}
        \toprule
        & Algorithm & $n = 100$ & $n=200$ & $n=300$  & $n=1000$\\
          \toprule
        \multirow{2}{5em}{RMSE} & $\MNN$ & $8.96 \pm 0.18$	& $8.93 \pm 0.06$ &   $\mathbf{8.44 \pm 0.12}$ &  $\mathbf{6.16 \pm 0.06}$	\\ \cline{2-6}
          & \SNN & $\mathbf{7.60 \pm 0.17}$	& $\mathbf{7.98 \pm 0.07}$  & -- & --	 \\
          \toprule
        \multirow{2}{5em}{MAE}	& $\MNN$	& $7.10 \pm 0.23$  & $7.06 \pm 0.05$ & $\mathbf{6.63 \pm 0.14}$  & $\mathbf{4.38 \pm 0.05}$	 \\ \cline{2-6}
        & \SNN & $\mathbf{4.53 \pm 0.08}$	& $\mathbf{4.78 \pm 0.03}$	&  -- & --  \\
          \toprule
\multirow{2}{5em}{Run Time} & $\MNN$ & $\le$ 10 s  & $\le$ 10 s & $\le$ 10 s & $\le$ 20 s \\ \cline{2-6}
& \SNN & 1 min 43 s  & 	40 min 19 s &  $\ge$ 24 h & $\ge$ 24 h \\
\bottomrule 
    \end{tabular}\label{table:exp-snn}
    \end{table}

\subsection{Practical implementation of $\MNN$} \label{sec:exp-mnn-implementation}

As highlighted before, in the experiments, we utilize a practical implementation of the proposed $\MNN$ algorithm where (i) $\bfA$ is used in place of $\tilde\bfA$ in the distance estimation step and (ii) the matrix completion from the clustered outcome matrix is done using SVD with known rank. Using such a practical implementation of the algorithm acts a proof of concept -- letting us determine the added value of the extra information gained from the two stage approach.

\paragraph{Regarding clustering in (i).} Typically one may perform clustering by either specifying the number of clusters or specifying a threshold distance as described in the proposed $\MNN$ in Section \ref{section:alg}. We use the subroutine the routine \href{https://scikit-learn.org/stable/modules/generated/sklearn.cluster.AgglomerativeClustering.html}{\texttt{AgglomerativeClustering}} from \texttt{scikit.cluster} which allows us to perform clustering in either ways. However, we still need to specify the number of clusters or the threshold distance that both depend on $\varepsilon_n$. In view of our discussion on the coverage property (see Appendix \ref{app:clustering}), a conservative choice on the number of clusters is $(3/\varepsilon_n)^{d}$, while a conservative choice of a distance threshold is $6 \varepsilon_n)$ as per the description of $\MNN$ in Section \ref{Alg:mnn}. Note that according to Theorem \ref{thm:prediction}, $\varepsilon_n = 2n^{-(2-\beta)/(d-1)}$ and the value of $\beta$ may not be known, but we may simply estimate as was the case for the Glance dataset. Practically, we observed that both ways of clustering perform similarly, but we had to tune the constants in front of $\varepsilon_n$ appropriately. In particular, when we performed clustering based on the number of clusters, we selected the number of cluster within the values $\lbrace  \sqrt{n}/2, \sqrt{n}, 2\sqrt{n}, 5\sqrt{n} \rbrace$, and when we performed clustering based on a distance threshold, we selected the thresholds within the values $\lbrace \varepsilon_n/5, \varepsilon_n/6, \varepsilon_n/7, \dots \varepsilon_n/90 \rbrace$. The selection process was done via cross-validation.

\paragraph{Regarding imputation in (ii).} Note that the only part of the proposed $\MNN$ that may be practically inefficient is the imputation subroutine as it requires to search over the space of  subsets of $[n]$ of size $d$, and such space is of size ${n \choose d}$. Thus, the imputation subroutine may have a computational complexity that is of order $n^{d\log(d)}$ which is still polynomial in $n$ but may not be practical for large values of $n$. Devising a practical efficient procedure to perform imputation under our model while enjoying entry-wise guarantees is an interesting open question that we leave for future work. For practical purposes, we recommend to use a standard matrix completion algorithm for this second phase such as USVT, ALS or nuclear norm penalization based estimators as we do. However, note that these methods are not proven to enjoy entry-wise guarantees under our setting which we find to be an interesting question that we leave for future work.

\section{Conclusion}

We proposed $\MNN$ a matrix completion method suitable for MNAR data models where there is shared information between the outcome and observation pattern. The proposed method provably exploits this shared information to improve predictions. The desirable theoretical properties and empirical performance of $\MNN$ highlight the value of modeling these shared latent factors. From a statistical learning perspective, we have shown that the parametric rate for supervised learning algorithms such as nearest-neighbor-style method can almost be achieved even when (i) covariates are unobserved, and (ii) observations are sampled in a biased manner; this could be of independent interest. Finally, we note that valuable future directions of research include (i) devising practical matrix completion methods for generic MNAR data models that are as statistically efficient as $\MNN$, and (ii) potentially relaxing conditional independence assumption between different entries of the observation pattern.

\newpage 
\bibliographystyle{alpha}
\bibliography{refs}

\appendix 
\newpage

\section{Proof of Proposition \ref{prop:clustering-denoising}}\label{sec:proof-prop-clustering-denoising}

\subsection{Distance Estimation}\label{app:estimation}
To guarantee good clustering performance, we must first show that the estimated distances converge to the true distances between pairs of users/items.
\begin{lemma}[Distance estimation] \label{lemma:distance} For all $\delta \in (0,1)$, we have 
        \begin{align}
            \PP\left(\max_{i \neq j } \left\vert \hatdu_{ij} -  \tilde{d}^\bu_{ij} \right\vert  \leq \frac{c}{\rho^2_n} \sqrt{\frac{d^3}{n} \log\parens*{\frac{e \, n}{\delta}}}
             \right) \ge 1 - \delta,
        \end{align}
        provided that $n \ge C d \log(e n /\delta)$, for some universal constants $c, C > 0$.
\end{lemma}

\paragraph{Proof of Lemma \ref{lemma:distance}.} Let $i,j \in [n]$, and let us recall that
        \begin{align} \label{eq:estimator}
            \hatdu_{ij} & = \frac{2}{\rho_n}\sqrt{\frac{d}{n}}~ \enorm{\widehat{\widetilde{\bfP}}_{i,:} - \widehat{\widetilde{\bfP}}_{j,:}}, \nonumber\\  
            \du_{ij} & = \Vert \bu_{i} - \bu_j \Vert_2. 
        \end{align}
        For ease of notations, let us further introduce the distance corresponding to the case where we know exactly the true observation probability matrix $\bfP$, 
        \begin{align}
            \tilde{d}^\bu_{ij} = \frac{2}{\rho_n}\sqrt{\frac{d}{n}}~ \enorm{\widetilde{\bfP}_{i,:} - {\widetilde{\bfP}}_{j,:}}.
        \end{align}
        Note that we can decompose the estimation error as follows using the triangular inequality
        \begin{align}
            \left\vert \hatdu_{ij} - \du_{ij} \right\vert \le \left\vert  \hatdu_{ij}  - \tilde {d}^\bu_{ij}\right\vert +  \left\vert  \tilde{d}^\bu_{ij} - \du_{ij} \right\vert. 
        \end{align}
        We will next proceed by bounding each term separately with high probability. 

        \medskip
        
        \underline{\emph{(1) Bounding $\vert  \tilde{d}^\bu_{ij} - \du_{ij}\vert$.}} We start by noting that 
        \begin{align*}
            \tilde{d}_{ij}^{\bu} =  \frac{2}{\rho_n}\sqrt{\frac{d}{n}}~ \enorm{\widetilde{\bfP}_{i,:} - \widetilde{\bfP}_{j,:}} &= \sqrt{\frac{d}{n}}~\enorm{\bfV(\bu_i - \bu_j)}.
        \end{align*}
        Thus, 
        \begin{align*}
            & \sqrt{\frac{d}{n}} \sigma_{\min}\parens*{\bfV} \enorm{\bu_i - \bu_j}  \leq \frac{2}{\rho_n}\sqrt{\frac{d}{n}}~ \enorm{\widetilde{\bfP}_{i,:} - \widetilde{\bfP}_{j,:}} \\
            & \qquad  \qquad \qquad \qquad \quad  \leq  \sqrt{\frac{d}{n}}  \sigma_{\max}\parens*{\bfV} \enorm{\bu_i - \bu_j},
        \end{align*}
        which entails that 
        \begin{align*}
            \left \vert \tilde{d}_{ij}^{\bu} - \du_{ij} \right \vert  & = \left \vert \frac{2}{\rho_n}\sqrt{\frac{d}{n}}~ \enorm{\widetilde{\bfP}_{i,:} - \widetilde{\bfP}_{j,:}}  - \Vert   \bu_i - \bu_j  \Vert_2  \right\vert \\
            & \le \max_{k \in [d]}  \left\vert    \sqrt{\frac{d}{n}}~ \sigma_k\left( \bfV \right) - 1 \right \vert   \Vert   \bu_i - \bu_j  \Vert_2 
        \end{align*}
        where $\sigma_{\min}(\bfX)$ (resp. $\sigma_{\max}(\bfX)$) refers to the smallest (resp. largest) singular value of $\bfX$. Since $\sqrt{d}\cdot\bfV$ is a random matrix with i.i.d. isotropic rows, using Lemma \ref{lemma:rowsv} and picking $t = \sqrt{\log(2/\delta)/c}$, we obtain that 
        \begin{align}\label{eq:svbound}
             \max_{k \in [d]} \abs*{ \sqrt{\frac{d}{n}}~\sigma_k\parens*{\bfV} - 1} \leq C\sqrt{\frac{d}{2 n} \log\parens*{\frac{2}{\delta}}}  
        \end{align}
        holds with probability at least $1 - \delta$, for some universal constant $C> 0$. Thus, recalling the simple fact $\Vert \bu_i - \bu_j\Vert_2 \le 2$, we conclude that 
        \begin{align} \label{eq:oracleerror}
            \PP\left(\left \vert \tilde{d}_{ij}^{\bu} - \du_{ij} \right \vert \leq C\sqrt{\frac{d}{n} \log\parens*{\frac{2}{\delta}}} \right) \ge 1 -\delta. 
        \end{align}

        \medskip 
        
        \underline{\emph{(2) Bounding $\vert \hatdu_{ij} -  \tilde{d}^\bu_{ij} \vert$.}} Using first the reverse triangular inequality, then the triangular inequality, we can further decompose the error term $\vert \hatdu_{ij} -  \tilde{d}^\bu_{ij} \vert$ as follows
        \begin{align}
            \left\vert \hatdu_{ij} -  \tilde{d}^\bu_{ij} \right\vert & \le \frac{2}{\rho_n}\sqrt{\frac{d}{n}} \left\Vert (\widehat{\widetilde{\bfP}}_{i,:} - \widehat{\widetilde{\bfP}}_{j,:} ) -  (\widetilde{\bfP}_{i,:} - \widetilde{\bfP}_{j,:} )\right\Vert_2    \\ 
            & \le  \frac{2}{\rho_n} \sqrt{\frac{d}{n}} \left(\left\Vert (\widehat{\widetilde{\bfP}}_{i,:} - \widetilde{\bfP}_{i,:}  \right \Vert + \left \Vert    \widehat{\widetilde{\bfP}}_{j,:}  - \widetilde{\bfP}_{j,:} \right\Vert_2  \right)
        \end{align}
        Therefore controlling the term $\vert \hatdu_{ij} -  \tilde{d}^\bu_{ij} \vert$ amounts to showing that $\Vert    \widehat{\widetilde{\bfP}}_{i,:}  - \widetilde{\bfP}_{i,:} \Vert_2$ is small for all $i \in [n]$. To that end, we use Lemma \ref{lemma:hsvt} to write 
        \begin{align}\label{eq:row-decomp}
            & \left \Vert \widehat{\widetilde{\bfP}}_{i,:}  - \widetilde{\bfP}_{i,:} \right\Vert_2^2 \le \nonumber\\
            & \quad 2 \left( \frac{\Vert \bfE \Vert_\op^2}{\sigma_d^2(\widetilde{\bfP})} \left( \Vert \bfE_{j, :} \Vert_2^2 + \Vert \widetilde{\bfP}_{j,:}\Vert_2^2    \right) + \Vert \bfR_d \bfR_d^\top \bfE_{j,:}\Vert_2^2   \right),
        \end{align}
        where we denote $\bfE = \widetilde{\bfA} -  \widetilde{\bfP} = \bfA - \bfP$, and  $\bfQ_d \bfSigma_d \bfR_d^\top$ the reduced SVD of $\widetilde{\bfP}$. Now, we will produce several bounds on the norms of the relevant matrices appearing in  \eqref{eq:row-decomp} separately.

        \medskip

        \emph{(2.1) Upper bounding $\opnorm{\bfE}$.} The entries of the noise matrix $\bfE$ are independent centered Bernoulli random variables, thus sub-gaussian. Using Lemma \ref{lemma:iidopnorm} and choosing $t = \sqrt{\log(2/\delta)/c}$ again, we obtain that the following 
        \begin{align} \label{eq:noiseopnorm}
            \opnorm{\bfE} &\leq 2C \sqrt{n} + \sqrt{\frac{\log(2/\delta)}{c}} \leq C' \sqrt{n \log(2/\delta)}.
        \end{align}
        holds with probability at least $1- \delta$, for some universal constants $c, C, C' > 0$.

        \medskip

        \emph{(2.2) Lower bounding $\sigma_d(\widetilde{\bfP})$.} We recall $\widetilde{\bfP} = (\rho_n / 2)  \bfU \bfV^\top$. By \eqref{eq:svbound}, the following  
        \begin{align}
            \sigma_d(\bfV) &\geq \sqrt{\frac{n}{d}} - C' \sqrt{\log{\parens*{\frac{2}{\delta}}}} \\
            &\geq C''\sqrt{\frac{n}{d}}.
        \end{align}
        holds with probability $1- \delta$, provided that $n \ge 
 (2 C')^2 d  \log(2/\delta)$ for some universal constants $C',C'' > 0$.
        By the same argument, we obtain a similar high probability lower bound on $\sigma_d(\bfU)$. Then, by a simple union bound, we obtain that 
        \begin{align} \label{eq:Pminsv}
            \sigma_d(\widetilde{\bfP} ) &\geq \frac{\rho_n}{2} \cdot \sigma_d(\bfU) \cdot \sigma_d(\bfV) \\
            &\geq \frac{C \rho_n n}{d},
        \end{align}
        holds with probability at least $1 - \delta$,  provided $n \ge c d \log(2/\delta)$, for some universal constants $c, C>0$.
        
        \medskip 

        \emph{(2.3) Upper bounding $\enorm{\bfE_{j, :}}$.} Note that $\Vert \bfE_{j,:} \Vert_2  \le \Vert \bfE \Vert_{\op}$. Thus, we have 
        \begin{equation} \label{eq:Ejnorm}
            \enorm{\bfE_{j,:}} \leq  C'\sqrt{n\log(2/\delta)}
        \end{equation}
        holds with probability at least $1 -\delta$, 
        as shown in \eqref{eq:noiseopnorm}.

        \medskip 

        \emph{(2.4) Upper bounding $\enorm{\widetilde{\bfP}_{j,:}}$.} Recall that $\widetilde{\bfP}_{j,:} = (\rho_n/2) \bfV \bu_j$. Using \eqref{eq:svbound}, which bounds the largest singular value of $\bfV$, we have that
        \begin{align} \label{eq:Pjnorm}
           \enorm{\widetilde{\bfP}_{j,:}} &= \frac{\rho_n}{2} \enorm{\bfV \bu_j} \\
           &\leq \frac{\rho_n}{2} \opnorm{\bfV} \enorm{\bu_j} \\
           &\leq \frac{\rho_n}{2} \parens*{\sqrt{\frac{n}{d}} + C' \sqrt{\log{\parens*{\frac{2}{\delta}}}}} \\
            &\leq \frac{C'' \rho_n}{2} \sqrt{\frac{n}{d}}.
        \end{align}
        holds with probability at least $1-\delta$, provided $n \ge c'd \log(2/\delta)$. 

        \medskip 

        \emph{(2.5) Upper bounding $\enorm{\bfR_d \bfR_d^\top \bfE_{j,:}}$.} First, since $\bfR_d$ contains orthonormal columns,
        \begin{align}
            \enorm{\bfR_d \bfR_d^\top \bfE_{j,:}} &\leq \opnorm{\bfR_d} \enorm{\bfR_d^\top \bfE_{j,:}} \\
            &\leq \enorm{\bfR_d^\top \bfE_{j,:}}.
        \end{align}
        We will bound this quantity using Lemma \ref{lemma:subgvec}. Towards showing that $\bfR_d^\top \bfE_{j,:}$ is a sub-gaussian random vector, consider any fixed $\bv \in \cS^{d-1}$. Then, denoting $\bfR_d = [r_{ki}]_{k \in [n], i \in [d]}$,
        \begin{equation}
            \bv^\top \bfR_d^\top \bfE_{j,:} = \sum_{i=1}^d \sum_{k=1}^n v_i r_{ki} \eta_{jk}
        \end{equation}
        which is a linear combination of independent sub-gaussian random variables $\eta_{jk} \sim \subG(1/4)$, and thus sub-gaussian. Moreover, its variance proxy is
        \begin{align}
            \frac{1}{4} \sum_{i=1}^d \sum_{k=1}^n v_i^2 r_{ki}^2 &= \frac{1}{4} \sum_{i=1}^d \parens*{v_i^2 \sum_{k=1}^n r_{ki}^2} \\
            &= \frac{1}{4} \sum_{i=1}^d v_i^2 \\
            &= \frac{1}{4}.
        \end{align}
        Therefore, $\bfR_d^\top \bfE_{j,:}$ is a $(1/4)$-sub-gaussian random vector. Using Lemma \ref{lemma:subgvec} with $t = \sqrt{\frac{d}{2} \log\parens*{\frac{2n}{\delta}}}$ and union bounding over all rows of $\bfE$ yields
        \begin{equation} \label{eq:projectnorm}
            \max_{j \in [n]}~\enorm{\bfR_d^\top \bfE_{j,:}} \leq \sqrt{\frac{d}{2} \log\parens*{\frac{2n}{\delta}}}.
        \end{equation}
        with probability at least $1-\delta$.

        \medskip 

        \emph{(2.6) Putting everything together.} Starting from \eqref{eq:row-decomp}, and using the bounds obtained in \eqref{eq:noiseopnorm}, \eqref{eq:Pminsv}, \eqref{eq:Ejnorm}, \eqref{eq:Pjnorm}, and \eqref{eq:projectnorm}, we have by substitution that
        \begin{align*}
            & \enorm{\widehat{\widetilde{\bfP}}_{i,:} - \widetilde{\bfP}_{i,:}}^2 \\
            & \quad \leq C \cdot \parens*{\frac{n}{\rho_n^2 n^2 / d^2} \parens*{C' n + \frac{\rho_n^2 n}{4d}} + \frac{d}{2} \log\parens*{\frac{2n}{\delta}}} \\
            &\quad\leq C \cdot \parens*{\frac{d^2}{\rho_n^2} \parens*{C' + \frac{\rho_n^2}{4d}} + \frac{d}{2} \log\parens*{\frac{2n}{\delta}}} \\
            &\quad\leq C \cdot \parens*{\frac{C' d^2}{\rho_n^2} + \frac{d}{2} \log\parens*{\frac{2n}{\delta}}} \\
            &\quad\leq \frac{C d^2}{\rho_n^2} \log\parens*{\frac{2n}{\delta}}.
        \end{align*}
        Scaling appropriately according to \eqref{eq:estimator}, and using the union bound, then yields
        \begin{equation} \label{eq:Phaterror}
            \PP\left(\left\vert \hatdu_{ij} -  \tilde{d}^\bu_{ij} \right\vert 
            \le \frac{C}{\rho^2_n} \sqrt{\frac{d^3}{n} \log\parens*{\frac{2n}{\delta}}} \right) \ge 1 - 10 \delta
        \end{equation}
         provided $n \ge c d \log(2/\delta)$ for some universal constants $c, C> 0$.

        \medskip 

        \emph{(3) Conclusion.} Finally, combining \eqref{eq:oracleerror} and \eqref{eq:Phaterror} with further simplifications gives 
        \begin{align}
            \PP\left(\max_{i \neq j } \left\vert \hatdu_{ij} -  \tilde{d}^\bu_{ij} \right\vert  \leq \frac{C}{\rho^2_n} \sqrt{\frac{d^3}{n} \log\parens*{\frac{e n}{\delta}}}
             \right) \ge 1 - \delta,
        \end{align}
        provided $n \ge c d \log(e n /\delta)$, for some universal constants $c, C > 0$.

\subsection{Clustering}\label{app:clustering}

\subsubsection{Coverage.} A key phenomenon that underpins our results is dense coverage of the latent feature space by users and items. Towards guaranteeing this, let $\Sigma_n \subset \cS^{d-1}$ be a minimal $\varepsilon_n$-net of $\cS^{d-1}$, with the properties that $\enorm{\bz - \bz'} > \varepsilon_n$ for distinct $\bz, \bz' \in \Sigma_n$ and for any $\bx \in \cS^{d-1}$ there exists $\bz \in \Sigma_n$ such that $\enorm{\bx - \bz} \leq \varepsilon_n$. It is widely known that $|\Sigma_n| \leq (1 + 2/\varepsilon_n)^d \leq (3 / \varepsilon_n)^d$ where the latter inequality assumes $\varepsilon_n \leq 1$, which we enforce.

\begin{lemma}[Coverage] \label{lemma:coverage}
Denote the $\varepsilon_n$-ball centered at $\bz$ as $\cB_{\varepsilon_n}(\bz)$, and let $(\bu_i)_{i \in [n]}$ be independent and uniformly random points on $\cS^{d-1}$. Then, for all $\delta \in (0,1)$,  
\begin{align}
    \PP \Big( \min_{\bz \in \Sigma_n} \sum_{i=1}^n \indicator \lbrace  \bu_i \in  \cB_{\varepsilon_n}(\bz)  \rbrace \ge \frac{n}{8} {\Big(\frac{\varepsilon_n}{2}\Big)}^{d-1} \Big) & \ge 1 - \delta,
\end{align}
provided that $\varepsilon_n$ satisfies 
\begin{align}
     n \left(\frac{\varepsilon_n}{2}\right)^{d-1} \ge C\log\left(\frac{e n}{\delta}\right), 
\end{align}
where $C$ is a positive universal constant. 
\end{lemma}

\paragraph{Proof of Lemma \ref{lemma:coverage}.} First, let us fix $\bz \in \Sigma_n$. Note that by assumption $\bu_1, \dots, \bu_n$ are i.i.d. on the unit sphere, so and an immediate application of multiplicative Chernoff bound with ratio 1/2 gives 

\begin{align}
    & \PP\Big(\sum_{i=1}^n \indicator\lbrace  \bu_i \in  \cB_{\varepsilon_n}(\bz)  \rbrace  \le \frac{n}{2}\,\PP(\bu_1 \in \cB_{\varepsilon_n}(\bz)) \Big) \nonumber \\ 
    & \qquad \leq \exp\Big( - \frac{n}{8} \, \PP(\bu_1 \in \mathcal{B}_{\varepsilon_n}(\bz)) \Big).
\end{align}
Now, by using Lemma \ref{lemma:caparea} we have 
    \begin{equation}
        \PP(\bu_i \in \cB_{\varepsilon_n}(\bz)) \geq \frac{1}{4} \sin^{d-1}\Phi
    \end{equation}
    where, using the geometry of the problem and recalling that $\varepsilon_n \leq 1$,
    \begin{equation*}
        \sin\Phi = \varepsilon_n \sqrt{1 - \varepsilon_n^2 / 4} \geq \varepsilon_n / 2.
    \end{equation*}
    Thus, we can write
    \begin{equation*}
        \PP(\bu_i \in \cB_{\varepsilon_n}(\bz)) \geq \frac{1}{4} \parens*{\frac{\varepsilon_n}{2}}^{d-1}.
    \end{equation*}
    This entails that 
    \begin{align*}
    & \PP\left(\sum_{i=1}^n \indicator\lbrace  \bu_i \in  \cB_{   \varepsilon_n}(\bz)  \rbrace  \le \frac{n}{8}  \left( \frac{\varepsilon_n}{2}\right)^{d-1} \right) \\
    & \qquad \qquad \qquad \quad \quad \le \exp\left( - \frac{n}{32} \, \parens*{\frac{\varepsilon_n}{2}}^{d-1} \right).
    \end{align*}
    Finally, by using an union bound over the $\varepsilon_n$-net $\Sigma_n$, we conclude that 
    \begin{align*}
        & \PP\left(  \min_{\bz \in \Sigma_n} \sum_{i=1}^n \indicator\lbrace  \bu_i \in  \cB_{   \varepsilon_n}(\bz)  \rbrace   \le \frac{n}{8}  \left( \frac{\varepsilon_n}{2}\right)^{d-1} \right) \\
        & \qquad\qquad\qquad\quad\quad\le \left(  \frac{3}{\varepsilon_n}  \right)^{d} \exp\left( - \frac{n}{32} \, \parens*{\frac{\varepsilon_n}{2}}^{d-1} \right) \\
        & \qquad\qquad\qquad\quad\quad\le \delta,
    \end{align*}
    where the last inequality is true whenever the condition  
    $
        n \left( \varepsilon_n/2\right)^{d-1} \ge 32\left(\log\left( (1 / \delta) \right) + d \log\left(3 / \varepsilon_n \right)\right), 
    $ holds, 
    which we may replace by the more conservative condition \begin{align}
            n \left( \frac{\varepsilon_n}{2}\right)^{d-1} \ge 32\log\left( \frac{e n }{64 \delta} \right).
    \end{align}

\subsubsection{Observations} Given that the latent feature space is densely covered by both users and items, we now argue that all pairs of user and item clusters -- as defined by $\Sigma_n$, rather than the central users and items -- that are reasonably close will generate a guaranteed number of observations.
\begin{lemma}[Observations] \label{lemma:observation}
     Consider $\bz_i, \bz_j \in \Sigma_n$ for which $\enorm{\bz_i - \bz_j} \leq \sqrt{2} + c \varepsilon_n$ for some absolute constant $c>0$, where we assume that $\varepsilon_n < (\sqrt{3} - \sqrt{2})/ (2 + c)$, and let the set of observations between these user and item clusters be denoted by 
    $\cY_{\varepsilon_n}^{(\bz_i, \bz_j)}  = \{y_{kl}: a_{kl} = 1, \bu_k \in \cB_{\varepsilon_n}(\bz_i), \bv_l \in \cB_{\varepsilon_n}(\bz_j)\}$.
    Then, for all $\delta \in (0,1)$, the event 
    \begin{equation*} \min_{\substack{\bz_i, \bz_j \in \Sigma_n: \\  \Vert \bz_i - \bz_j \Vert_2 \le \sqrt{2} + c \varepsilon_n }} \abs*{\cY_{\varepsilon_n}^{(\bz_i, \bz_j)}} \geq c n^2 \rho _n  \left( \frac{\varepsilon_n}{2}\right)^{2d-2},
    \end{equation*}
    holds with probability at least $1- \delta$, 
    as long as $\varepsilon_n$ satisfies 
    \begin{align*}
        \left(\frac{\varepsilon_n}{2}\right)^{d-1} \ge \frac{C}{n}\left (\sqrt{\frac{1}{\rho_n}\log\left( \frac{e \, n}{\delta} \right) } + \log\left( \frac{e \, n}{\delta}\right) \right)
    \end{align*}
    for some universal constants $c, C> 0$.
\end{lemma}

\paragraph{Proof of Lemma \ref{lemma:observation}.}
        Let us define the events 
        \begin{align*}
            \cE_1 & = \left\lbrace \sum_{k=1}^n \indicator \lbrace \bu_k \in \cB_{\varepsilon_n}(\bz_i)\rbrace \ge \frac{n}{8}\left( \frac{\varepsilon_n}{2}\right)^{d-1} \right\rbrace, \\
            \cE_2 & = \left\lbrace \sum_{\ell=1}^n \indicator \lbrace \bv_\ell \in \cB_{\varepsilon_n}(\bz_j)\rbrace  \ge \frac{n}{8}\left( \frac{\varepsilon_n}{2}\right)^{d-1} \right\rbrace, 
        \end{align*}
        and note that according to Lemma \ref{lemma:coverage} and union bounding over the two events we have 
        \begin{align*}
            \PP(\cE_1 \cap \cE_2 ) \ge 1 - 2 \delta
        \end{align*}
        provided that $ 1 \ge \varepsilon_n \ge 2 \left( \frac{C}{n} \log\left(\frac{e n}{\delta}\right)  \right)^{\frac{1}{d-1}}$. Next, observe that for any $\bu_k \in \cB_{\varepsilon_n}(\bz_i), \bv_l \in \cB_{\varepsilon_n}(\bz_j)$ where $\enorm{\bz_i - \bz_j} \leq \sqrt{2} - 2 \varepsilon_n$, we have
        \begin{align*}
            \enorm{\bu_k - \bv_l} & \leq \enorm{\bz_i - \bz_j} + \enorm{\bu_k - \bz_i} + \enorm{\bv_l - \bz_j} \\
            & \leq \sqrt{2} + (2 + c)\varepsilon_n \\
            & \le \sqrt{3}.
        \end{align*}
        Thus,
        \begin{align}
            p_{kl} &= \frac{\rho_n}{2} (\bu_k^\top \bv_l + 1) \\
            &= \frac{\rho_n}{2} \parens*{2 - \frac{1}{2} \enorm{\bu_k - \bv_l}^2} \\
            &\geq \rho_n / 4.
        \end{align}
        Now, conditioned on the event $\cE_1 \cap \cE_2$, we know that there are at least $\frac{n}{8}  \left( \frac{\varepsilon_n}{2}\right)^{d-1}$ users and items in the balls around $\bz_i$ and $\bz_j$ respectively, all pairs of which satisfy the above inequality. Defining 
        \begin{align*}
            \cE_3 = \left\lbrace  \min_{\substack{\bz_i, \bz_j \in \Sigma_n: \\  \Vert \bz_i - \bz_j \Vert_2 \le \sqrt{2} + c \varepsilon_n }}  \abs*{\cY_{\varepsilon_n}^{(\bz_i, \bz_j)}} \ge \frac{n^2 \rho _n}{512} \left( \frac{\varepsilon_n}{2}\right)^{2d-2} \right\rbrace, 
        \end{align*}
        then using a multiplicative Chernoff bound and union bounding over all pairs of points in $\Sigma_n$, we get
        \begin{align*}
            \PP\left(\cE_3^c \vert \cE_1 \cap \cE_2  \right) & \leq \parens*{\frac{3}{\varepsilon_n}}^{2d} \exp\parens*{-\frac{\rho_n n^2}{2048} \left( \frac{\varepsilon_n}{2}\right)^{2d-2}} \\
            & \le \delta,
        \end{align*}
        where the last inequality holds as long as 
        \begin{align*}
        \rho_n n^2 \left(\frac{\varepsilon_n}{2} \right)^{2d-2} \ge 2048 \left( \log\left(\frac{1}{\delta}\right) + 2d \log\left(\frac{3}{\varepsilon_n}\right)\right).
        \end{align*}
        Note that $n (\varepsilon_n/2)^{d-1} \ge C' \log(e n / \delta)$ ensures that $\log(n)  \ge d \log(\varepsilon_n/3)$ for $C'$ larger than $1$. Thus, the above condition can be enforced if, 
        \begin{align}
            \left(\frac{\varepsilon_n}{2}\right)^{d-1} \ge \frac{C}{n}\left (\sqrt{\frac{1}{\rho_n}\log\left( \frac{e \, n}{\delta} \right) } + \log\left( \frac{e \, n}{\delta}\right) \right)
        \end{align}
        for some positive universal constants $C$ that are large enough. Finally, we conclude by noting that 
        \begin{align*}
            & \PP\left(\min_{\substack{\bz_i, \bz_j \in \Sigma_n: \\  \Vert \bz_i - \bz_j \Vert_2 \le \sqrt{2} + c \varepsilon_n }}\abs*{\cY_{\varepsilon_n}^{(\bz_i, \bz_j)}} < \frac{\rho_n n^2}{512} \left( \frac{\varepsilon_n}{2}\right)^{2d-2}  \right)  \\
            & \qquad \qquad\qquad   \le \PP\left(\cE_3^c \vert \cE_1 \cap \cE_2  \right) + \PP(\cE_1^c \cup \cE_2^c)  \\
            & \qquad \qquad\qquad \le 3 \delta.
        \end{align*}     

\subsection{Proof of Proposition \ref{prop:clustering-denoising}} \label{app:denoising}
Now we are ready to prove Proposition \ref{prop:clustering-denoising}. To start with, we can express the difference between $\bar h_{ij}$ and $h_{ij}$ for any $i \in \cU^\star, j \in \cV^\star$ as $\bar h_{ij} - h_{ij}  = \Delta_{ij} + \nu_{ij}$, where 
\begin{align*}
    \Delta_{ij} & = \frac{1}{|\Omega_{ij}|} \sum_{(k, l) \in \Omega_{ij}} \left(f(\bu_k, \bv_l) - f(\bu_i, \bv_j)\right)
    \\
    \nu_{ij} & = \frac{1}{|\Omega_{ij}|} \sum_{(k, l) \in \Omega_{ij}} \xi_{kl}
\end{align*}
        \underline{\emph{(1) Bounding $\nu_{ij}$.}} Recall that an entry of $\overline{\bfH}$ is considered revealed if $\abs{\Omega_{ij}} \geq N_n = c \,n^2 \rho_n \left(\varepsilon_n/2\right)^{2d-2}$. Thus, $\nu_{ij}$ is the mean of \emph{at least} $N_n$ independent subgaussian random variables of variance proxy $\sigma^2$, and by union bounding Lemma \ref{lemma:subgmean} over all $n^2$ user-item pairs we have that the event:
        \begin{equation}
            \max_{(i, j): \bar h_{ij} \neq \star} \abs{\nu_{ij}} \geq \sqrt{\frac{\sigma^2}{2c n^2 \rho_n} \left(\frac{2}{\varepsilon_n}\right)^{2d-2} \log\left(\frac{2n^2}{\delta} \right)},
        \end{equation}
        holds with probability at most $\delta$
        \medskip 

        \underline{\emph{(2) Bounding $\Delta_{ij}$.}} Moreover, by Lemma \ref{lemma:distance}, all the estimated user-user and item-item distances have error bounded by $\varepsilon_n$ if 
        \begin{align*}
            \ddist(n) \triangleq \frac{C}{\rho_n^2}\sqrt{\frac{d^3}{n} \log\left( \frac{e \, n}{\delta}\right)} \le \varepsilon_n.
        \end{align*}
        Since for any $k \in \cU_i$ it must be true that $\hatdu_{ik} \leq 6\varepsilon_n$ by construction (otherwise user $k$ itself would have been a central user), we know the true distance is also small, i.e. $\enorm{\bu_i - \bu_k} \leq \hatdu_{ik} + \ddist(n) \leq 7\varepsilon_n$. The same holds for $\enorm{\bv_j - \bv_l}$ for any $l \in \cV_j$. Thus, by the Lipschitz property of $f$, every $(k, l) \in \Omega_{ij}$ satisfies
        \begin{equation*}
            \abs*{f( \bu_k,  \bv_l) - f(\bu_i, \bv_j)} \leq 14 L B \varepsilon_n.
        \end{equation*}
        Combining the two bounds yields the desired bound.

\newpage 
\section{Proof of Proposition \ref{prop:imputation-1}}\label{app:imputation-1}

Once  clustering $\&$ denoising have been executed, the resulting matrix $\overline{\bfH}$ has still some missing entries, and its low-rank structure is a priori unclear. In Proposition \ref{prop:imputation-1}, we argue that $\overline{\bfH}$ has rank at least $d$, and describe the pattern of its revealed entries. The theoretical tools we use to prove Proposition \ref{prop:imputation-1} exploit the connection between the geometry of the problem and the low-rank structure present in our model. These tools are presented in Section \ref{app:proxy-features}.

\subsection{Proof of Proposition \ref{prop:imputation-1}.} 

To start with, let us present an important result about the existence of unit vectors with some crucial properties that will prove useful for the proof of Proposition \ref{prop:imputation-1}.

\begin{proposition}\label{prop:existence-of-proxy-feature} For all $\bu, \bv \in \cS^{d-1}$, there exists $\bx_1, \dots, \bx_d, \by_1, \dots, \by_d \in \cS^{d-1}$, such that: 
    \begin{itemize}
        \item [(i)] for all $i,j \in [d]$, $\Vert \bv - \bx_i \Vert_2 \le \sqrt{2}$, $\Vert \bu - \by_j \Vert_2 \le \sqrt{2}$ and $\Vert \bx_i - \by_j\Vert_2 \le \sqrt{2}$,
        \item [(ii)] for all $i,j \in [d]$, $i \neq j$, we have $\Vert \bx_i - \bx_j\Vert_2 \wedge \Vert \by_i - \by_j\Vert_2 \ge L^{-1} \kappa^2 \sqrt{(1-\cos(\pi/12)) \cos(\pi/12)}$,
        \item [(iii)] and the following holds 
        $$
        \sigma_d\left(\begin{bmatrix} \theta(\bx_1) & \dots & \theta(\bx_d) \end{bmatrix}^\top \begin{bmatrix}
            \phi(\by_1) & \hdots & \phi(\by_d) 
        \end{bmatrix} \right)  \ge \frac{m^2 \kappa^4 (1- \cos(\pi/12)) \cos(\pi/12)  }{2}
        $$
    \end{itemize}
\end{proposition}

We postpone the proof of Proposition \ref{prop:existence-of-proxy-feature} until the next section.  With this we are now ready to present the proof of Proposition \ref{prop:existence-of-proxy-feature}. Let $i \in \cU^\star,j\in \cV^\star$ such that $\bar h_{ij} = \star$, and recall that  $h_{ij} = f(\bu_i, \bv_j)$. We start by invoking Proposition \ref{prop:existence-of-proxy-feature}, which ensures, following Algorithm \ref{Alg:proxy-features}, that there exists $\bx_1, \dots, \bx_d,$ and $ \by_1, \dots, \by_d $ in $\cS^{d-1}$, such that: 
\begin{itemize}
        \item [\emph{(i)}] For all $k,\ell \in [d]$, $\Vert \bv_j - \bx_k \Vert_2 \le \sqrt{2}$, $\Vert \bu_i - \by_\ell \Vert_2 \le \sqrt{2}$ and $\Vert \bx_k - \by_\ell \Vert_2 \le \sqrt{2}$;
        \item [\emph{(ii)}] For all $i,j \in [d]$, $i \neq j$, we have $\Vert \bx_i - \bx_j\Vert_2 \wedge \Vert \by_i - \by_j\Vert_2 \ge L^{-1} \kappa^2 \sqrt{(1-\cos(\pi/12)) \cos(\pi/12)}$;
        \item [\emph{(iii)}] And the following holds 
        $$
        \sigma_d\left(\begin{bmatrix} \theta(\bx_1) & \dots & \theta(\bx_d) \end{bmatrix}^\top \begin{bmatrix}
            \phi(\by_1) & \hdots & \phi(\by_d) 
        \end{bmatrix} \right)  \ge \frac{m^2 \kappa^4 (1- \cos(\pi/12)) \cos(\pi/12)  }{2}
        $$
\end{itemize}
Next, we argue that there exists $\cI = \lbrace i_1, \dots, i_d \rbrace \subseteq  \cU^\star$ and $\cJ = \lbrace j_1, \dots, j_d \rbrace \subseteq \cV^\star$ such that 
\begin{align}\label{eq:within-dist}
    \Vert \bx_k - \bu_{i_k} \Vert_2 \le 8 \varepsilon_n \quad \text{and} \quad \Vert \by_\ell - \bv_{j_\ell} \Vert_2 \le 8 \varepsilon_n. 
\end{align}
Indeed, recall that $\Sigma_n$ is an $\epsilon_n$-net, thus  there exists $\bz_1, \dots, \bz_d \in \Sigma_n$ such that $\Vert \bz_k - \bx_k  \Vert_2 \le \varepsilon_n$. Moreover thanks to property $(ii)$, satisfied by $\bx_1, \dots, \bx_d$, the vectors $\bz_1, \dots, \bz_d$ are distinct, with every pair being  $14 \varepsilon_n$ apart if 
\begin{align}
    14 \varepsilon_n < \frac{\kappa^2}{L} \sqrt{(1- \cos(\pi/12) ) \cos(\pi/12)} 
\end{align}
Therefore, using Lemma \ref{lemma:coverage}, there must exist $i_1', \dots, i_d' \in [n]$ such that $\Vert \bu_{i_k'} - \bz_k \Vert \le \varepsilon_n$ for all $k \in [d]$, which in turn ensures existence of $i_1, \dots, i_d \in \cU^\star$ such that $\Vert \bu_{i_k} - \bz_k \Vert \le 7\varepsilon_n$. Finally, using property \emph{(i)} together with triangular inequality ensures \eqref{eq:within-dist}. Existence of $j_1, \dots, j_k \in \cV^\star$ satisfying \eqref{eq:within-dist} is proven similarly. 

\medskip

Consequently, we obtain that for all $k, \ell \in [d]$, 
\begin{align}\label{eq:observation}
    \Vert \bu_i - \bv_{j_\ell} \Vert_2 & \le \sqrt{2} + 8\varepsilon_n, \nonumber \\
    \Vert \bu_{i_k} - \bv_{j_\ell} \Vert_2 & \le \sqrt{2} + 16\varepsilon_n, \nonumber \\ \quad \Vert \bu_{i_k} - \bv_{j} \Vert_2 
    & \le \sqrt{2} + 8\varepsilon_n
\end{align}
and by Lemma \ref{lem:perturbation} together with property $(iii)$, we obtain 
\begin{align}
    \sigma_d\left( \bfH_{\cI, \cJ} \right) & = \sigma_d\left( \left(\theta(\bu_{i_k})^\top \phi(\bv_{j_\ell})\right)_{k,\ell \in [d]} \right) \\
    & \ge \frac{m^2 \kappa^4 (1- \cos(\pi/12)) \cos(\pi/12)  }{4}
\end{align}
provided the following condition holds
\begin{align}
    8 \varepsilon_n <  \frac{m^2 \kappa^4 (1- \cos(\pi/12)) \cos(\pi/12) }{8 BL}
\end{align}

Since $\Sigma_n$ is an $\varepsilon_n$-net, using  using \eqref{eq:observation}, there exists $\bz_i, \bz_{i_1}, \dots, \bz_{i_d}$ and $\bz_j, \bz_{j_1}, \dots, \bz_{j_d}$ in $\Sigma_n$ such that  for all $k,\ell \in [d]$, we have 
\begin{align*}
    & \bu_i \in \cB_{\varepsilon_n}(\bz_i), \quad \bu_{i_k} \in \cB_{\varepsilon_n}(\bz_{i_k}), \\
    &  \bv_j \in \cB_{\varepsilon_n}(\bz_j),  \quad \bv_{i_\ell} \in \cB_{\varepsilon_n}(\bz_{i_\ell})
\end{align*}
and 
\begin{align*}
    \Vert \bz_i - \bz_{j_\ell} \Vert_2 & \le \sqrt{2} + 10\varepsilon_n, \\
    \quad  \Vert \bz_{i_k} - \bz_{j_\ell} \Vert_2 & \le \sqrt{2} + 18\varepsilon_n, \\
    \quad \Vert \bz_{i_k} - \bz_{j} \Vert_2 & \le \sqrt{2} + 10\varepsilon_n,
\end{align*}
thus, given the choice of $N_n$, we may apply Lemma \ref{lemma:observation} to conclude that the event: 
\begin{align*}
    \forall i' \in \cI, \forall j' \in \cJ, \ \ \bar{h}_{ij'} \neq \star, \ \ \bar{h}_{i'j'} \neq \star, \ \ \bar{h}_{i'j} \neq \star, 
\end{align*}
holds with probability at least $1- \delta$, provided that $\varepsilon_n < (\sqrt{3} - \sqrt{2})/20$ and 
\begin{align}
    \left(\frac{\varepsilon_n}{2}\right)^{d-1} \ge \frac{C}{n}\left (\sqrt{\frac{1}{\rho_n}\log\left( \frac{e \, n}{\delta} \right) } + \log\left( \frac{e \, n}{\delta}\right) \right)
\end{align}
This concludes the proof.

\subsection{Proof of Proposition \ref{prop:existence-of-proxy-feature}}\label{app:proxy-features}

\subsubsection{Properties of $\theta$ and $\phi$}

We present a couple of properties of $\theta$ and $\phi$ that are key in establishing our result. 
\begin{lemma}\label{lem:homeomorphism}
    If $\theta(\cdot)$ is non-degenerate with parameter $m,\kappa >0$, then the following properties hold:
    \begin{itemize}
        \item [(i)] The map $\theta$ is an homeomorphism from $\cS^{d-1}$ to $\theta(\cS^{d-1})$. In fact, it is an $(L, m\kappa)$-bilipschitz map. 
        \item [(ii)] The map $\tilde{\theta}$ is an homeomorphism from $\cS^{d-1}$ to $\cS^{d-1}$. In fact, it is an $(L/m, \kappa)$-bilipchitz map.
    \end{itemize} 
\end{lemma}

\paragraph{Proof of Lemma \ref{lem:homeomorphism}.}
    To prove  \emph{(i)} and \emph{(ii)}, it suffices to establish that the mappings $\theta$ and $\tilde{\theta}$ are bi-Lipchitz with the appropriate constants, which in turn ensures the weaker statement that both mappings are homeomorphisms.

    We start by showing \emph{(i)}. Let $\bx, \by \in \cS^{d-1}$, and observe that we have the following elementary identity:
   \begin{align}\label{eq:elem}
               \Vert \theta(\bx) - \theta(\by)\Vert^2 = e_1 + e_2    
   \end{align}
   with $e_1 = \left\vert \Vert \theta(\bx) \Vert - \Vert \theta(\by)\Vert  \right\vert^2$ and $e_2 = \Vert\theta(\bx) \Vert \Vert \theta(\by) \Vert \left\Vert \tilde{\theta}(\bx) - \tilde{\theta}(\by)\right\Vert^2$.
    Given that $\theta(\cdot)$ is non-degenerate with parameters $m,\kappa >0$, it holds that 
    \begin{align*}
        \Vert \theta(\bx) - \theta(\by)\Vert^2 & \ge  \Vert\theta(\bx) \Vert \Vert \theta(\by) \Vert \left\Vert \tilde{\theta}(\bx) - \tilde{\theta}(\by)\right\Vert^2 \\
        & \ge m^2 \kappa^2 \Vert \bx - \by \Vert^2.      
    \end{align*}
    Thus, it must hold that $\theta$ is $(L, m\kappa)$-bilipschitz. Consequently, $\theta$ is injective and continuous and its inverse is also continuous. Thus, it is an homeomorphism.

    \medskip 

    Now, we prove \emph{(ii)}. First, note that the non-degeneracy property of $\theta$ ensures that $\tilde \theta$ is well defined. In addition to that, and using \eqref{eq:elem} we have for all $\bx, \by \in \cS^{d-1}$,
    \begin{align*}
        L^2 \Vert \bx - \by \Vert^2 & \ge \Vert \theta(\bx) - \theta(\by)\Vert^2, \\
        \Vert \theta(\bx) - \theta(\by)\Vert^2 & \ge m^2 \Vert \tilde{\theta}(\bx) - \tilde \theta(\by) \Vert^2
    \end{align*} 
    Thus, it follows that $\tilde{\theta}$ is $(L/m, \kappa)$-bi-Lipchitz. This means that $\tilde \theta$ is continuous on $\cS^{d-1}$, is bijection from $\cS^{d-1}$ to $\tilde \theta(\cS^{d-1}) \subseteq \cS^{d-1}$, and its inverse is also continuous. This means that $\tilde \theta$ is a homeomorphsim and it must be the case that $\tilde\theta(\cS^{d-1}) = \cS^{d-1}$.

\begin{lemma}\label{lem:cap-to-cap}
    Let $\cC$ be a spherical cap on the unit sphere $\cS^{d-1}$ defined by the point $\bx$, and polar angle $\alpha$. Then, $\tilde{\theta}(\cC)$ contains the spherical cap defined by the point $\tilde{\theta}(\bx)$, and polar angle at least $\tilde \alpha = \arccos(1-\kappa^2 + \kappa^2 \cos(\alpha))$.  
\end{lemma}

\paragraph{Proof of Lemma \ref{lem:cap-to-cap}.}  
First, let us define $\tilde{\cC} \subseteq \tilde{\theta}(\cC)$, to be a spherical determined by the point $\tilde{\theta}(\bx)$ and maximal polar angle $\tilde{\alpha}$ (by maximal we mean that any spherical cap that is defined by the point $\tilde{\theta}(\bx)$ and  has polar angle $\tilde{\alpha} + \varepsilon$ is no more included in $\tilde{\theta}(\cC)$ for any $\varepsilon > 0$). 

We claim that $\tilde \alpha \ge \arccos(1-\kappa^2 + \kappa^2 \cos(\alpha) )$. Indeed, since $\tilde{\cC} $ has maximal polar angle it must be the case that the boundary of $\tilde\cC$, denoted $\partial \tilde{\cC}$, intersects the boundary $\partial (\tilde{\theta}(\cC))$ at least at a point $z$. Moreover, since $\tilde{\theta}$ is a homeomorphism we have  $\partial \tilde{\theta}(\cC) = \tilde{\theta}(\partial \cC)$, thus $\bz \in \partial \tilde\cC \cap \tilde{\theta}(\partial \cC)$ and there must exist $\bx_z \in \partial \cC$ such that $\tilde{\theta}(\bx_\bz) = \bz$. Therefore, we have 
\begin{align*}
      2 - 2 \cos(\tilde \alpha)  = \Vert \tilde\theta(\bx) - \bz \Vert^2 = \Vert \tilde\theta(\bx) - \tilde\theta(\bx_\bz), \Vert^2 
\end{align*}
and 
\begin{align*}
      \Vert \tilde\theta(\bx) - \tilde\theta(\bx_\bz) \Vert^2 \ge \kappa^2 \Vert \bx - \bx_\bz\Vert^2 = \kappa^2 (2 - 2 \cos(\alpha)).
\end{align*}
Rearranging the terms in the above inequality we obtain 
\begin{align*}
    \cos(\tilde \alpha) \le (1 - \kappa^2) + \kappa^2 \cos(\alpha), 
\end{align*}
which leads to $\tilde \alpha \ge \arccos(1-\kappa^2 + \kappa^2 \cos(\alpha))$ (because $\cos(\cdot)$ is a strictly decreasing function from $[0,\pi]$ to $[-1, 1]$, and $\alpha, \tilde\alpha \in [0, \pi] $).

\subsubsection{Properties of high-dimensional simplices}

The following result on regular simplicies is borrowed from \cite{parks2002elementary} (see their Theorem 1). 

\begin{lemma} \label{lem:dihedral-angle-regular-simplex}
   Let $\ba_1, \dots, \ba_d$ be unit vectors such that $\lbrace \ba_1, \dots, \ba_d \rbrace$ form an $(d-1)$-regular simplex embedded in $\mathbb{R}^{d}$. Then, we have for all $i,j \in [d]$ such that $i\neq j$, we have  
   \begin{align*}
       \langle \ba_i, \ba_j \rangle = \frac{-1}{d-1}.
   \end{align*}
\end{lemma}

 Below, we present a result on the singular values of a matrix formed by a set of linearly independent vectors that are equidistant.   

\begin{lemma}\label{lem:non-trivial-singular-value}
    Let $\bx_1, \dots, \bx_d \in \cS^{d-1}$ be unit vectors that are equidistant, i.e. there exists a constant $c>0$, such that for all $i,j\in [d]$, $i \neq j$, $\Vert \bx_i - \bx_j \Vert = c$, then, denoting $\bfX = \begin{bmatrix} \bx_1 & \cdots & \bx_d\end{bmatrix}$, we have 
    \begin{align*}
        \sigma_1(\bfX) & = \sqrt{1 + (d-1)\left(1- \frac{c^2}{2}\right)},  \\
        \quad \sigma_2 (\bfX) & = \dots = \sigma_d (\bX) = \frac{c}{\sqrt{2}}. 
    \end{align*}
\end{lemma}

\paragraph{Proof of Lemma \ref{lem:non-trivial-singular-value}.} Let us start by observing that, for all $i,j \in [d]$, 
    \begin{align*}
        (\bfX^\top \bfX)_{ij} = \begin{cases}
            1 & \text{ if } i = j, \\
            (2 -  c^2)/2 & \text{ if } i \neq j.
        \end{cases}
    \end{align*}
    where we used the fact that $(\bfX^\top \bfX)_{ij} = \bx_i^\top \bx_j = (\Vert \bx_i \Vert^2 + \Vert \bx_j\Vert^2 - \Vert \bx_i - \bx_j \Vert^2)/2$. Next, we will provide $\bz_1, \dots, \bz_d$,  $d$ orthogonal eigenvectors of $\bfX^\top \bfX$, which will allow us to explicitely provide the singular values of $\bfX$. We define
    \begin{align*}
        \bz_1^\top  & = \begin{bmatrix}
            1 
            & 1 
            & 1 & 
            \cdots 
            & 1
        \end{bmatrix}, \\
        \bz_2^\top & = \begin{bmatrix}
            -1 
            & d-1 
            & -1 & 
            \cdots 
            & -1
        \end{bmatrix}, \\
        \bz_3^\top & = \begin{bmatrix}
            -1 
            & -1 
            & d-1 & 
            \cdots 
            & -1
        \end{bmatrix}, \\
        &  \vdots \\
        \bz_d^\top  & = \begin{bmatrix}
            -1 
            & -1 
            & -1 & 
            \cdots 
            & d-1
        \end{bmatrix}, \\
    \end{align*}
    It is not difficult to verify that $\bz_1, \dots, \bz_d$ are linearly independent. Furthermore, we can also verify that 
    \begin{align*}
        (\bfX^\top \bfX) \bz_1 = (1 + (d-1)(2-c^2)/2) \bz_1, 
    \end{align*}   
    and for all $i \in \lbrace 2, \dots, d\rbrace$, we have 
    \begin{align*}
        (\bfX^\top \bfX) \bz_i =  (c^2/2)  \bz_i.
    \end{align*}
    This proves that $\bfX^\top \bfX$ has two eigenvalues such that the smallest one $(c^2/2)$ has multiplicity $d-1$. We conclude that 
    $
        \sigma_1(\bfX) = \sqrt{(1 + (d-1)(2-c^2)/2)}, \quad \text{and} \quad \sigma_2(\bfX) = \cdots = \sigma_d(\bfX) = c/\sqrt{2} 
    $

\subsubsection{Proof of Proposition \ref{prop:existence-of-proxy-feature}} We prove existence of $\bx_1, \dots, \bx_d, \by_1, \dots, \by_d$ satisfying properties $(i)$ and $(ii)$ by construction. This construction is geometric in nature and draws inspiration from properties of high dimensional simplices. We provide an algorithmic description of such construction in Algorithm \ref{Alg:proxy-features}. \\

\begin{algorithm}[ht!] 
    \caption{Search rank $d$-sub-matrix in $\bfX$}
    \label{Alg:proxy-features}
    \begin{algorithmic}
    \State \textbf{Input:} $\bu, \bv \in \cS^{d-1}$\;
    \State \emph{\color{purple}\underline{Step 1: Finding sufficiently close features}} 
    \State Find the geodesic $\Gamma$ between $\bu$ and $\bv$ on the hypersphere $\cS^{d-1}$\;
     \State Find the midpoint $\bfm \in \cS^{d-1}$ of $\Gamma$ \; 
    \State Find the point $\bx$ on $\Gamma$ such that $\arccos (\langle \bx, \bfm \rangle) = \pi/8$ and is closer to $\bv$
    \;
     \State  Find the point $\by$ on $\Gamma$ such that $\arccos (\langle \by, \bfm \rangle) = \pi/8$ and is closer to $\bu$
    \;
    \State Let $\cC(\bx, \pi/12)$ be the spherical cap on $\cS^{d-1}$ specified by the point $\bx$ and polar angle $\pi/12$\;
    \State Let $\cC(\by, \pi/12)$ be the spherical cap on $\cS^{d-1}$ specified by the point $\by$ and polar angle $\pi/12$\;
    \State \emph{\color{purple}\underline{Step 2: Finding well separated features}} 
    \State Let $\alpha = \arccos (1-\kappa^2 + \kappa^2 \cos(\pi/12))$\;
    \State Let $\cC(\tilde\theta(\bx), \alpha )$ be the spherical cap on $\cS^{d-1}$ specified by the point $\tilde\theta(\bx)$ and polar angle $\alpha$\;
    \State Let $\cC(\tilde\phi(\by), \alpha)$ be the spherical cap on $\cS^{d-1}$ specified by the point $\tilde\phi(\by)$ and polar angle $\alpha$\;
    \State Let $\bz_1, \dots, \bz_d$ be vertices of a regular $(d-1)$-simplex inscribed in $\partial\cC(\tilde\theta(\bx), \alpha)$\;
    \State Let $\bw_1, \dots, \bw_d$ be vertices of a regular $(d-1)$-simplex inscribed in $\partial\cC(\tilde\phi(\by), \alpha)$\; 
    \State For all $i \in [d]$, set $\bx_i = \tilde{\theta}^{-1}(\bz_i)$ and $\by_i = \tilde \phi^{-1}(\bw_i)$\;
    \State \textbf{Output:} $\bx_1, \dots, \bx_d, \by_1, \dots, \by_d \in \cS^{d-1}$
    \end{algorithmic}
\end{algorithm}

    \underline{\textbf{Claim 1.}} Step 1 in Algorithm \ref{Alg:proxy-features}, ensures that any choice of points $\bx_1, \dots, \bx_d \in \cC(\bx, \pi/12)$ and $\by_1, \dots, \by_d  \in \cC(\by, \pi/12)$ satisfies property \emph{(i)}.  \\ 

    \emph{Proof of Claim 1.} Let $\bx' \in \cC(\bx, \pi/12)$ and $\by' \in \cC(\by, \pi/12)$. Note that by construction $\langle \bx, \bv  \rangle \ge \cos(\pi/2 - \pi/8)$ (because $\bfm, \bx, \bv$ are coplanar and $\langle\bx, \bfm \rangle \ge \cos(\pi/2)$). Let $\bx' \in \cC(\bx, \pi/12)$. Then, we have $\langle \bv, \bx'\rangle \ge \cos( 3\pi/8 + \pi/12) \ge 0$. This entails that $\Vert \bx' - \bv \Vert \le \sqrt{2}$. We can prove similarly that for all $\by' \in \cC(\by, \pi/12)$, we have $\Vert \by' - \bu \Vert \le \sqrt{2}$. Next, note also that $\langle \bx', \by'  \rangle \ge \cos(2\pi/12 + 2 \pi/8) \ge 0$, thus $\Vert \bx' - \by' \Vert \le \sqrt{2}$. This concludes the proof of the claim. \\

    \underline{\textbf{Claim 2.}} Step 2 in Algorithm \ref{Alg:proxy-features}, ensures that the selected vectors $\bx_1, \dots, \bx_d$ and $\by_1, \dots, \by_d$ satisfy property \emph{(iii)}. \\ 

    \emph{Proof of Claim 3.} First, we note that Lemma \ref{lem:homeomorphism} and Lemma \ref{lem:cap-to-cap} ensure that $\tilde{\theta}$ and $\tilde \phi$ are well defined,  and that $\cC(\tilde{\theta}(\bx), \alpha) \subseteq \tilde{\theta}(\cC(\bx, \pi/12))$ and $\cC(\tilde{\phi}(\by), \alpha) \subseteq \tilde\phi(\cC(\by, \pi/12))$. This in turn implies that $\bx_1, \dots, \bx_d \in \cC(\bx, \pi/12)$ and $\by_1, \dots, \by_d \in \cC(\by, \pi/12)$. It remains to lower bound the the smallest singular value of $\begin{bmatrix}
        \theta(\bx_1) & \hdots & \theta(\bx_d)
    \end{bmatrix}^\top  \begin{bmatrix}
        \phi(\by_1) & \hdots & \phi(\by_d)
    \end{bmatrix}$. Under assumption \textbf{A2}, we have 
    \begin{align*}
        & \sigma_d \left(\begin{bmatrix}
            \theta(\bx_1) & \hdots & \theta(\bx_d)
        \end{bmatrix} \right)^2  =  \inf_{\bz \in \cS^{d-1}} \sum_{i =1 }^d \vert \theta(\bx_i)^\top \bz \vert^2 \\
        & \qquad \qquad \qquad \qquad  \ge m^2 \inf_{\bz \in \cS^{d-1}} \sum_{i =1 }^d \vert \tilde\theta(\bx_i)^\top \bz \vert^2 \\
        & \qquad \qquad \qquad \qquad \ge m^2 \sigma_d\left( \begin{bmatrix}
            \tilde \theta(\bx_1) & \hdots & \tilde \theta(\bx_d)
        \end{bmatrix}\right)^2 
    \end{align*}
    where we used the variational definition of the smallest singular value. Now, observe that by construction $\tilde\theta(\bx_1), \dots,  \tilde\theta(\bx_d)$ are the vertices of an $(d-1)$-simplex inscribed in $\partial \cC(\tilde{\phi}, \alpha)$. Thus, for each $i \neq j$, we have 
    \begin{align*}
        \left\Vert \tilde\theta(\bx_i) - \tilde\theta(\bx_j)\right\Vert =  \sqrt{1 - \cos^2(\alpha)}. 
    \end{align*}
    Therefore, we may apply Lemma \ref{lem:non-trivial-singular-value} and obtain: 
    \begin{align*}
        \sigma_d \left(\begin{bmatrix}
            \tilde \theta(\bx_1) & \hdots & \tilde \theta(\bx_d)
        \end{bmatrix} \right)^2 \ge \frac{1 - \cos^2(\alpha)}{2}
    \end{align*}
    We can similarly lower bound $\sigma_d \left(\begin{bmatrix}
        \tilde \phi(\by_1) & \hdots & \tilde \phi(\by_d)
    \end{bmatrix} \right)$. We conclude that 
    \begin{align*}
        & \sigma_d\left(\begin{bmatrix}
            \theta(\bx_1) & \hdots & \theta(\bx_d)
        \end{bmatrix}^\top  \begin{bmatrix}
            \phi(\by_1) & \hdots & \phi(\by_d)
        \end{bmatrix} \right) \\
        & \qquad \qquad \ge \frac{m^2 (1 - \cos^2(\alpha))}{\sqrt{2}} \\
        & \qquad \qquad= \frac{m^2\left(1 -  \left(1-\kappa^2 + \kappa^2 \cos(\pi/12) \right)^2\right) }{\sqrt{2}} \\
        & \qquad \qquad \ge \frac{m^2 \kappa^4 (1- \cos(\pi/12)) \cos(\pi/12)  }{\sqrt{2}}.
    \end{align*}
    
    \medskip 
    
    \underline{\textbf{Claim 3.}} Step 2 in Algorithm \ref{Alg:proxy-features} also ensures that the selected vectors $\bx_1, \dots, \bx_d, \by_1, \dots, \by_d$ satisfy property $(ii)$. \\

    Indeed, for all $i \neq j$, we have by construction that $\Vert \tilde{\theta}(\bx_i) - \tilde{\theta}(\bx_j)\Vert_2 = \sqrt{1-\cos^2(\alpha)}$. Then, using the fact that $\widetilde{\theta}(\cdot)$ is $L$-lipschitz which comes from Lemma \ref{lem:homeomorphism}, we can immediately lower bound $\Vert \bx_i - \bx_j \Vert_2$ as claimed. The claim for  $\by_1, \dots, \by_d$ follows similarly.

\subsubsection{Smallest singular value under perturbation}

The Proposition \ref{prop:existence-of-proxy-feature} guarantees existence of $\bx_1, \dots, \bx_d$ and $\by_1, \dots, \by_d$ in $\cS^{d-1}$ such that
$\sigma_d\left(\begin{bmatrix} \theta(\bx_1) & \hdots & \theta(\bx_d) \end{bmatrix}\right) \ge C $ and $ 
    \sigma_d\left(\begin{bmatrix}
        \phi(\by_1) & \dots & \phi(\by_d)
    \end{bmatrix} \right) \ge C $
for some universal positive constant $C> 0$. However, we may only have approximate values of the vectors $\bx_1, \dots, \bx_d$ and $\by_1,\dots, \by_d$. Below, we present a result showing that this does not matter too much. 

\begin{lemma}\label{lem:perturbation}
    Let $\bx_1, \dots, \bx_d$ and $\by_1, \dots, \by_d$ in $\cS^{d-1}$ such that 
    $
        \sigma_d\left( \theta(\bfX)^\top \phi(\bfY)\right) \ge C.
    $
    Let $\bu_1, \dots, \bu_d$ and $\bv_1, \dots, \bv_d \in \cS^{d - 1}$ be such that  
    $
        \Vert \bU - \bfX\Vert_{2 \to \infty} \le \varepsilon$ and   $\Vert \bV - \bfY\Vert_{2 \to \infty} \le  \varepsilon 
    $
    for some $\varepsilon < C/(4BL)$. Then, we have 
\begin{align*}
     \sigma_d\left( \theta(\bU)^\top \phi(\bV)\right) \ge \frac{C}{2}
\end{align*}
\end{lemma}

\paragraph{Proof of Lemma \ref{lem:perturbation}}
As $\theta$ and $\phi$ are $L$-Lipschitz, we know
\begin{align*}
    \abs*{\theta(\bu_i)^\top\phi(\bv_j) - \theta(\bx_i)^\top\phi(\by_j)} \leq 2BL\varepsilon
\end{align*}
Thus, applying Weyl's inequality, we have
\begin{align*}
    & \abs*{\sigma_d(\theta(\bU)^\top\phi(\bV)) - \sigma_d(\theta(\bX)^\top\phi(\bY))} \\
    & \qquad \qquad \leq \sigma_1(\sigma_d(\theta(\bU)^\top\phi(\bV)) - \theta(\bfX)^\top\phi(\bfY)) \\
    & \qquad \qquad \leq \Vert \theta(\bU_n)^\top \phi(\bfV_n) - \theta(\bfX)^\top \phi(\bfY) \Vert_{\max} \\
    &\qquad \qquad \leq \max_{i, j \in [d]} \abs*{\theta(\bu_i)^\top\phi(\bv_j) - \theta(\bx_i)^\top\phi(\by_j)} \\
    & \qquad \qquad \leq 2BL\varepsilon
\end{align*}
If $\varepsilon \leq C/(4BL)$, then we have
\begin{equation*}
    \sigma_d\parens*{\theta(\bU)^\top\phi(\bV)} \geq \frac{C}{2}
\end{equation*}

\newpage 
\section{Proof of Proposition \ref{prop:imputation2}} \label{app:imputation-2}

We are now ready to prove Proposition \ref{prop:imputation2}. Under the assumptions made in the proposition, we establish that the following event:   
    \begin{align*}
        & \max_{(i,j) \in \cU^\star \times \cV^\star }\vert \hat{h}_{ij} - h_{ij} \vert \le c \left(\frac{d^3B^2}{(m^2\kappa^4 \wedge m^4\kappa^8)} \wedge 1 \right) \\
        & \qquad \qquad \times \left( L B \varepsilon_n  + \frac{\sigma}{n} \left(\frac{2}{\varepsilon_n}\right)^{d-1} \sqrt{ \frac{1}{\rho_n}\log\left(\frac{e\, n}{\delta}\right)}  \right)
    \end{align*}
    holds with probability at least $1 - \delta$, provided that 
    \begin{align}
        \left( \frac{\varepsilon_n}{2}\right)^{d-1} & \ge  \frac{C d}{m^2 \kappa^4 n}\left (\sqrt{\frac{1}{\rho_n}\log\left( \frac{e \, n}{\delta} \right) } + \log\left( \frac{e \, n}{\delta}\right) \right) \\
        \varepsilon_n & \ge \frac{C}{\rho_n^2} \sqrt{\frac{d^3}{n} \log\left( \frac{e\, n}{\delta}\right) } \\
        \varepsilon_n & \le C \left( \frac{\kappa^4 (m^2 \wedge 1)}{d L(B\vee 1)} \wedge 1 \right), \\ 
          n & \ge C d \log\left(\frac{e \, n}{\delta}\right),
    \end{align}
    for some universal constants $c, C > 0$.

\paragraph{Proof of Proposition \ref{prop:imputation2}.}
    Fix $i \in \cU^\star,j \in \cV^\star$ such that $\hat{h}_{ij} = \star$. 
    
    \medskip 
    \underline{\emph{Step 1. (Revealed entries of interest).}}
    By Proposition \ref{prop:imputation-1}, the event: there exists $\cI \subseteq \cU^\star$, $\cJ \subseteq \cV^\star$ such that $\vert\cI \vert = \vert \cJ \vert = d$, and $\sigma_d(\bfH_{\cI, \cJ}) \ge c \, m^2 \kappa^4 $
    \begin{align}\label{eq:event1}
        \forall i' \in \cI, \forall j' \in \cJ, \ \ \bar{h}_{ij'} \neq \star, \ \ \bar{h}_{i'j'} \neq \star , \ \ \bar{h}_{i'j} \neq \star, 
    \end{align}
    holds with probability at least $1-\delta$, where $c>0$ is an absolute constant, provided $\varepsilon_n$ satisfies the condition \eqref{condition:eps-2}.

    \medskip 
    
    \underline{\emph{Step 2. (Error bound for revealed entries).}} Next, we know by Proposition \ref{prop:clustering-denoising}, that the event: 
    for all entries $(i, j) \in \cU^\star \times \cV^\star$, such that $\bar h_{ij} \neq \star$, 
    \begin{align}\label{eq:event 2}
        & \vert \bar{h}_{ij} - h_{ij} \vert \le \dnoise(n)  \nonumber \\
        &  \qquad \triangleq 14 L B \varepsilon_n +  \sqrt{\frac{\sigma^2}{2c n^2 \rho_n} \left(\frac{2}{\varepsilon_n}\right)^{2d-2} \! \! \!  \log\left(\frac{2n^2}{\delta} \right)}
    \end{align}
    holds with probability at least $1-\delta$, provided that $\varepsilon_n$ satisfies condition \eqref{condition:eps-1} and $n$ is such that $n \gtrsim d \log\left( e\,n /\delta \right)$.

    \medskip

    \underline{\emph{Step 3. (Error bound for imputed entries).}}
     Now, we are ready to use Lemma \ref{lemma:rlest}, which, under the intersection of the two events described by  \label{eq:event1} and \label{eq:event2}, entails that      
    \begin{align*}
        & \vert \hat{h}_{ij} - h_{ij} \vert \\
        & \  \le \left( \sqrt{2} + \frac{4\sqrt{2}d^2B^2}{\sigma_d(\bfH_{\cI, \cJ})} +  \frac{2(1 + \sqrt{5})4d^3B^2}{\sigma_d^2(\bfH_{\cI, \cJ})} \right) \dnoise(n) \\
        & \ \le \left( \sqrt{2} + \frac{4\sqrt{2}d^2B^2}{c m^2 \kappa^4 } +  \frac{2(1 + \sqrt{5})4d^3B^2}{c^2 m^4\kappa^8} \right) \dnoise(n)
    \end{align*}
    provided that the condition $
        \dnoise(n)d \le cm^2\kappa^4/4$ holds, which we may be enforced by assuming that 
    \begin{align*}
       \left( \frac{\varepsilon_n}{2}\right)^{d-1} & \ge \frac{C d}{m^2\kappa^4 n} \sqrt{\frac{1}{\rho_n} \log\left( \frac{e\, n}{\delta}\right)} \\
       \varepsilon_n & \le \frac{C m^2\kappa^4}{d L B} 
    \end{align*}
    for some universal constant $C> 0$ sufficiently large.

\newpage
\section{Helper Lemmas}
\begin{lemma} \label{lemma:caparea}
    Let the surface area of the unit $d$-sphere be $A_d$. Consider the corresponding hyperspherical cap subtended by the angle $\Phi \in (0, \pi/2)$, and let $A_d^{\operatorname{cap}}$ be its surface area. It holds that
    \begin{equation}
        A_d^{\operatorname{cap}}/A_d \geq \frac{1}{4} \sin^{d-1}\Phi.
    \end{equation}
\end{lemma}
    \paragraph{Proof of Lemma \ref{lemma:caparea}.}
        Using a result from \cite{li2010concise}, we write
        \begin{equation}
            A_d^{\operatorname{cap}}/A_d = \frac{1}{2} I \parens*{\sin^2\Phi; \frac{d-1}{2}, \frac{1}{2}},
        \end{equation}
        where $I(x; a, b)$ is known as the \emph{regularized incomplete beta function} and defined by
        \begin{equation}
            I(x; a, b) = \frac{B(x; a, b)}{B(a, b)}
        \end{equation}
        where $B(a, b)$ is the beta function and
        \begin{equation}
            B(x; a, b) = \int_0^x t^{a-1} (1-t)^{b-1} dt
        \end{equation}
        is the incomplete beta function. Then
        \begin{align}
            B\parens*{\sin^2\Phi; \frac{d-1}{2}, \frac{1}{2}} &= \int_0^{\sin^2\Phi} t^{\frac{d-3}{2}} (1 - t)^{-\frac{1}{2}} dt \\
            &\geq \int_0^{\sin^2\Phi} t^{\frac{d-3}{2}} dt \\
            &= \frac{2}{d-1} \sin^{d-1}\Phi.
        \end{align}
        Meanwhile, a simple but tight bound \cite{alzer2001sharp} for the denominator is
        \begin{align}
            B\parens*{\frac{d-1}{2}, \frac{1}{2}} \leq \frac{4}{d - 1}.
        \end{align}
        Combining the bounds yields
        \begin{equation}
            A_d^{\operatorname{cap}}/A_d \geq \frac{1}{4} \sin^{d-1}\Phi.
        \end{equation}

\begin{lemma}[Adapted from Appendix G.4 of \cite{agarwal2022multivariate}] \label{lemma:hsvt}
    Let $\bfY = \bfM + \bfE \in \RR^{m \times n}$ with $\mathrm{rank}(\bfM) = d$. Let $\bfU_d \bfSigma_d \bfV_d^\top$ and $\hat\bfU_d \hat\bfSigma_d \hat\bfV_d^\top$ denote the top $d$ singular components of the SVD of $\bfM$ and $\bfY$ respectively. Then, the truncated SVD estimator $\hat\bfM = \hat\bfU_d \hat\bfSigma_d \hat\bfV_d^\top$ is such that for all $j \in [m]$,
    \begin{align*}
        & \enorm{\hat \bfM_{j} - \bfM_{j}}^2 \leq \\
        & \qquad 2 \cdot \parens*{\frac{\opnorm{\bfE}^2}{\sigma_d^2(\bfM)} \parens*{\enorm{\bfE_{j}}^2 + \enorm{\bfM_{j}}^2} + \enorm{\bfV_d \bfV_d^\top \bfE_{j}}^2}.
    \end{align*}
    \end{lemma}
    \paragraph{Proof of Lemma \ref{lemma:hsvt}.}
        First note that $\hat\bfM = \bfY \hat\bfV_d \hat\bfV_d^\top$ by definition of the truncated SVD. Thus we can write
        \begin{align} 
            & \hat \bfM_{j} - \bfM_{j} \\
            & \quad = (\hat\bfV_d \hat\bfV_d^\top \bfY_{j} - \hat\bfV_d \hat\bfV_d^\top \bfM_{j}) + (\hat\bfV_d \hat\bfV_d^\top \bfM_{j} - \bfM_{j}) \\
            &\quad = \hat\bfV_d \hat\bfV_d^\top \bfE_{j} + (\hat\bfV_d \hat\bfV_d^\top - \bfI_n)~\bfM_{j}. \label{eq:hsvt1}
        \end{align}
        Since the two terms belong to orthogonal subspaces, we have
        \begin{equation} 
            \enorm{\hat \bfM_{j} - \bfM_{j}}^2 = \enorm{\hat\bfV_d \hat\bfV_d^\top \bfE_{j}}^2 + \enorm{(\hat\bfV_d \hat\bfV_d^\top - \bfI_n)~\bfM_{j}}^2. \label{eq:hsvt2}
        \end{equation}
        The first term of \eqref{eq:hsvt2} can be expanded by the triangle inequality as
        \begin{align}  
            & \enorm{\hat\bfV_d \hat\bfV_d^\top \bfE_{j}}^2 \\
            &\ \ \leq 2 \cdot \parens*{\enorm{\hat\bfV_d \hat\bfV_d^\top \bfE_{j} - \bfV_d \bfV_d^\top \bfE_{j}}^2 + \enorm{\bfV_d \bfV_d^\top \bfE_{j}}^2} \\
            &\ \ \leq 2 \cdot \parens*{\opnorm{\hat\bfV_d \hat\bfV_d^\top - \bfV_d \bfV_d^\top}^2 \enorm{\bfE_{j}}^2 + \enorm{\bfV_d \bfV_d^\top \bfE_{j}}^2}. \label{eq:hsvt3}
        \end{align}
        Since $\mathrm{rank}(\bfM) = d$, the Wedin $\sin{\Theta}$ Theorem \cite{wedin1972perturbation} guarantees that
        \begin{equation} 
            \opnorm{\hat\bfV_d \hat\bfV_d^\top - \bfV_d \bfV_d^\top} \leq \frac{\opnorm{\bfE}}{\sigma_d(\bfM)}. \label{eq:hsvt4}
        \end{equation}
        Combining \eqref{eq:hsvt3} and \eqref{eq:hsvt4} yields
        \begin{equation} 
            \enorm{\hat\bfV_d \hat\bfV_d^\top \bfE_{j}}^2 \leq 2 \cdot \parens*{\frac{\opnorm{\bfE}^2}{\sigma_d^2(\bfM)} \enorm{\bfE_{j}}^2 + \enorm{\bfV_d \bfV_d^\top \bfE_{j}}^2}. \label{eq:hsvt5}
        \end{equation}
        Now we return to the second term of \eqref{eq:hsvt2}. By definition, $\bfM = \bfM \bfV_d \bfV_d^\top$. Therefore, using \eqref{eq:hsvt4} once again, we get
        \begin{align} 
            \enorm{\hat\bfV_d \hat\bfV_d^\top \bfM_{j} - \bfM_{j}}^2 &= \enorm{(\hat\bfV_d \hat\bfV_d^\top - \bfV_d \bfV_d^\top)~\bfM_{j}}^2 \\
            &\leq \opnorm{\hat\bfV_d \hat\bfV_d^\top - \bfV_d \bfV_d^\top}^2 \enorm{\bfM_{j}}^2 \\
            &\leq \frac{\opnorm{\bfE}^2}{\sigma_d^2(\bfM)} \enorm{\bfM_{j}}^2. \label{eq:hsvt6}
        \end{align}
        Combining \eqref{eq:hsvt2}, \eqref{eq:hsvt5}, and \eqref{eq:hsvt6} we obtain the desired bound.

\begin{lemma}[Proposition 2.4 of \cite{rudelson2010non}] \label{lemma:iidopnorm}
    Let $\bfX = [x_{ij}] \in \Reals^{m \times n}$ be a random matrix whose entries are independent mean-zero sub-gaussian random variables with bounded sub-gaussian norm $\norm{x_{ij}}_{\psi_2} \leq 1$. Then for any $t \geq 0$, with probability at least $1 - 2\exp(-ct^2)$ it holds that
    \begin{equation}
        \opnorm{\bfX} \leq C(\sqrt{m} + \sqrt{n}) + t,
    \end{equation}
    where $C, c > 0$ are absolute constants.
\end{lemma}

\begin{lemma}[Matrix Bernstein inequality] \label{lemma:matrixbernstein}
    Let $\bfX_1,\dots,\bfX_n$ be a sequence of independent random symmetric matrices with dimension $d$. Assume that each matrix satisfies
    \begin{align*}
        \E\sqbrac{\bfX_i} &= 0, \\
        \lambda_{\max}(\bfX_i) &\leq R.
    \end{align*}
    Further define
    \begin{equation}
        \sigma^2 = \opnorm{\sum_{i=1}^n \E\sqbrac{\bfX_i^2}}.
    \end{equation}
    Then
    \begin{align*}
        & \prob{\lambda_{\max}\left(\sum_{i=1}^n \bfX_i\right) \geq t} \leq d \exp{\left(\frac{-t^2}{\sigma^2 + Rt/3}\right)} 
    \end{align*}
\end{lemma}

\begin{lemma} [Theorem 5.39 of \cite{vershynin2010introduction}] \label{lemma:rowsv}
    Let $\bfX \in \Reals^{n \times d}$, where $d < n$, have rows sampled i.i.d. from an isotropic distribution, i.e. $\E[\bfX_i \bfX_i^\top] = \bfI_d$. Then for every $t \geq 0$, with probability at least $1 - 2 \exp(-ct^2)$ the singular values of $\bfX$ satisfy
    \begin{equation*}
        \forall i \in [d], \ \sqrt{n} - C \sqrt{d} - t \leq \sigma_{i}(\bfX) \leq \sqrt{n} + C \sqrt{d} + t,
    \end{equation*}
    where $C, c > 0$ are constants that only depend on the sub-gaussian norm of $\bfX_i$, i.e. $\norm{\bfX_i}_{\psi_2}$.
\end{lemma}

\begin{lemma}[Mean of independent sub-gaussian random variables] \label{lemma:subgmean}
Let $X_1,\dots,X_n$ be a sequence of independent mean-zero random variables with $X_i \sim \subG(\sigma_i^2)$. Then, for any $t > 0$, it holds that
\begin{equation}
    \prob{\abs*{\frac{1}{n}\sum_{i=1}^n X_i} > t} \leq 2 \exp{\parens*{-\frac{n^2 t^2}{2 \sum_{i=1}^n \sigma_i^2}}}.
\end{equation}
\end{lemma}

\begin{lemma}[Norm of sub-gaussian random vector] \label{lemma:subgvec}
    Let $\bx \in \Reals^n$ be a mean-zero $\sigma^2$-sub-gaussian random vector, i.e. for any fixed $\bv \in \cS^{n - 1}$ the random variable $\bv^\top \bx$ is $\sigma^2$-sub-gaussian, or
    \begin{equation}
        \E[\exp(s \bv^\top \bx)] \leq \exp\parens*{\frac{s^2 \sigma^2}{2}}.
    \end{equation}
    Then, with probability at least $1 - 2\exp(-\frac{t^2}{2n\sigma^2})$, it holds that
    \begin{equation}
        \enorm{\bx} \leq t.
    \end{equation}
\end{lemma}

The proof of Lemma \ref{lemma:subgvec} is immediate. Since $\bx$ is $\sigma^2$-sub-gaussian, it is $(n\sigma^2)$-norm-sub-gaussian as defined \cite{jin2019short}. Using Definition 3 and Lemma 1 of \cite{jin2019short}, the result immediately follows.

\begin{lemma}[Lemma 12 of \cite{shahRL}]\label{lemma:schur}
    Let $M = \begin{bmatrix} A & B \\ C & D \end{bmatrix}$. If $\mathrm{rank}(A) = \mathrm{rank}(M)$, then $D = CA^\dagger B$.
\end{lemma}

\begin{lemma}[Theorem 3.3 of \cite{stewartinverse}]\label{lemma:perturb}
    For any matrices $A, B$ with $B = A + \Delta$, it holds that
    \begin{equation}
        \opnorm{B^\dagger - A^\dagger} \leq \frac{1 + \sqrt{5}}{2}\max \left(\opnorm{A^\dagger}^2, \opnorm{B^\dagger}^2 \!\right) \! \!\opnorm{\Delta}.
    \end{equation}
\end{lemma}

\begin{lemma}[Adapted from Proposition 13 of \cite{shahRL} ]\label{lemma:rlest}
Let $\bfH$ be an $\vert \cU^\star\vert \times \vert \cV^\star \vert$  dimensional matrix with entries are in $[-d B^2 , d B^2]$, and satisfying $\mathrm{rank}(\bfH) = d$. Let $\overline{\bfH}$ be an alternative $\vert \cU^\star\vert \times \vert \cV^\star \vert$  dimensional matrix. Now, for any $i \in \cU^\star, j \in \cV^\star$, assume that there exists $\cI \subseteq \cU^\star$ and $\cJ \subseteq \cV^\star$, such that $ \vert \cI \vert = \vert \cJ \vert = d$, and for all $k \in \cI, \ell \in \cJ$, 
\begin{align}
    \vert \bfH_{i\ell} - \overline{\bfH}_{i\ell} \vert \le \epsilon, \  \vert \bfH_{k\ell} - \overline{\bfH}_{k\ell} \vert \le \epsilon, \ \vert \bfH_{k j} - \overline{\bfH}_{k j}\vert  \le \epsilon,
\end{align}
for some  $0 < \epsilon \le \sigma_d(\bfH_{\cI,\cJ})/4d$. Then, defining 
\begin{align}
    \widehat{\bfH}_{ij} = \overline{\bfH}_{i,\cJ} \left(\overline{\bfH}_{\cI,\cJ}\right)^{\dagger} \overline{\bfH}_{\cI, j},
\end{align}
it holds that 
\begin{align}
     \abs*{\widehat{\bfH}_{i,j} - \bfH_{i,j}} \!\leq\!\left( \! \! \sqrt{2}\! + \!\frac{4\sqrt{2}d^2B^2}{\sigma_d(\bfH_{\cI, \cJ})}\! + \! \frac{2(1 + \sqrt{5})4d^3B^2}{\sigma_d^2(\bfH_{\cI, \cJ})} \! \right)\! \epsilon
\end{align}
\end{lemma}

\paragraph{Proof of Lemma \ref{lemma:rlest}.}
        First, observe that for any $ i \in \cU^\star, j  \in \cV^\star$, we have by Lemma $\ref{lemma:schur}$ that
        \begin{equation}\label{eq:truthschur}
            \bfH_{ij} = \bfH_{i, \cJ}\left(\bfH_{\cI, \cJ} \right)^\dagger\bfH_{\cI, j}
        \end{equation}
        as $\bfH_{\cI, \cJ}$ and $\bfH$ both have rank $d$.
        
        Next, we fix some $ i \in \cU^\star,  v\in \cV^\star$ and consider the error $\widehat \bfH_{ij} - \bfH_{ij}$. According to the definition of $\widehat \bfH$ and \eqref{eq:truthschur}, 
        \begin{align}
            & \widehat \bfH_{i,j} - \bfH_{i,j} \nonumber \\
            & \qquad =  \overline{\bfH}_{i, \cJ}\left(\overline{\bfH}_{\cI, \cJ} \right)^\dagger\overline{\bfH}_{\cI, j}  -  \bfH_{i, \cJ}\left(\bfH_{\cI, \cJ} \right)^\dagger\bfH_{\cI, j} \\
            & \qquad = \overline{\bfH}_{i, \cJ}\left(\overline{\bfH}_{\cI, \cJ} \right)^\dagger\overline{\bfH}_{\cI, j} -  \bfH_{i, \cJ}\left(\overline{\bfH}_{\cI, \cJ} \right)^\dagger\bfH_{\cI, j} \nonumber \\
            & \qquad \ \ \ + \bfH_{i, \cJ}\left(\overline{\bfH}_{\cI, \cJ} \right)^\dagger\bfH_{\cI, j} -  \bfH_{i, \cJ}\left(\bfH_{\cI, \cJ} \right)^\dagger\bfH_{\cI, j} \\
            & \qquad = \mathrm{Tr} \left( \left(\overline{\bfH}_{\cI,\cJ} \right)^{\dagger} \left( \overline{\bfH}_{ \cI, j} \overline{\bfH}_{i, \cJ}  - \bfH_{ \cI, j} \bfH_{i, \cJ} \right)  \right) \nonumber \\
            & \qquad \ \ \ + \mathrm{Tr} \left(  
\left((\overline{\bfH}_{\cI,\cJ})^\dagger   - ({\bfH}_{\cI,\cJ})^\dagger \right) \left( \bfH_{\cI, j} \bfH_{i, \cJ} \right)  \right)
        \end{align}
        Since $\abs{\mathrm{Tr} (AB)} \leq \sqrt{\mathrm{rank}(B)}\opnorm{A}\fnorm{B}$, we obtain
        \begin{align}
            & \abs*{\widehat{\bfH}_{i,j} - \bfH_{i,j}} \\
            &\quad \leq \sqrt{2}\opnorm{\left(\overline{\bfH}_{\cI, \cJ}\right)^\dagger} \fnorm{\overline{\bfH}_{\cI, j}  \overline{\bfH}_{i, \cJ} - {\bfH}_{\cI, j}  {\bfH}_{i, \cJ} } \label{eq:first}\\
            &\quad + \opnorm{(\overline{\bfH}_{\cI,\cJ})^\dagger   - ({\bfH}_{\cI,\cJ})^\dagger } \fnorm{\bfH_{\cI, j} \bfH_{i, \cJ}}. \label{eq:second}
        \end{align}
        Moving forward, we will separately establish upper bounds for the two terms. To upper bound \eqref{eq:first}, note first that by assumption $\opnorm{\Delta} \leq 2d\maxnorm{\Delta} \leq 2\epsilon d$, where $\Delta = \overline\bfH_{\cI, \cJ} - \bfH_{\cI,\cJ} $. Since $\sigma_d(\overline\bfH_{\cI, \cJ}) \geq \sigma_d(\overline\bfH_{\cI, \cJ}) - \opnorm{\Delta}$ by Weyl's inequality, we have
        \begin{align*}
            \opnorm{(\overline\bfH_{\cI, \cJ})^\dagger} & = \frac{1}{\sigma_d(\overline\bfH_{\cI, \cJ}) } \\
            & \leq \frac{1}{\sigma_d(\overline\bfH_{\cI, \cJ}) - 2\epsilon d} \\
            & \le \frac{2}{\sigma_d(\overline\bfH_{\cI, \cJ})}.
        \end{align*}
        To bound the second part of the expression, note $\fnorm{\overline{\bfH}_{\cI, j}  \overline{\bfH}_{i, \cJ} - {\bfH}_{\cI, j}  {\bfH}_{i, \cJ} } \leq 2d(2dB^2\epsilon + \epsilon^2)$ because $\fnorm{\overline{\bfH}_{i', j}  \overline{\bfH}_{i, j'} - {\bfH}_{i', j}  {\bfH}_{i, j'} } \leq 2dB^2\epsilon + \epsilon^2$ for all $(i',j') \in \cU^\star \times \cV^\star$.
        Now, to upper bound \eqref{eq:second}, we first upper bound $\opnorm{(\overline{\bfH}_{\cI,\cJ})^\dagger   - ({\bfH}_{\cI,\cJ})^\dagger }$ by using Lemma \ref{lemma:perturb} to note
        \begin{equation}
            \opnorm{(\overline{\bfH}_{\cI,\cJ})^\dagger   - ({\bfH}_{\cI,\cJ})^\dagger } \leq \frac{1 + \sqrt{5}}{2} \cdot \frac{8\epsilon d.}{\sigma_d^2(\bfH_{\cI, \cJ})} 
        \end{equation}
        Also it is easy to see that $\fnorm{\bfH_{\cI, j} \bfH_{i, \cJ}} \leq 2d^2B^2$. Combining all these bounds together with the condition that $\epsilon d \le \sigma_d(\bfH_{\cI, \cJ})/4$gives
        \begin{align}
            & \abs*{\widehat{\bfH}_{i,j} - \bfH_{i,j}} \\
            &\qquad \leq \left(\frac{4\sqrt{2}(d^2B^2 + d \epsilon) }{\sigma_d(\bfH_{\cI, \cJ}) } +  \frac{2(1 + \sqrt{5})4d^3B^2}{\sigma_d^2(\bfH_{\cI, \cJ})} \right) \epsilon \\
            & \qquad \le \left( \sqrt{2} + \frac{4\sqrt{2}d^2B^2}{\sigma_d(\bfH_{\cI, \cJ})} +  \frac{2(1 + \sqrt{5})4d^3B^2}{\sigma_d^2(\bfH_{\cI, \cJ})} \right) \epsilon
        \end{align}


\end{document}